%% file: main.tex
\documentclass[final,5p,times,twocolumn]{elsarticle}

\usepackage{amssymb}

\usepackage{amsmath,amsfonts,amsthm,graphicx}
\usepackage{booktabs}
\usepackage{comment}
\usepackage{color}
\usepackage{epstopdf}
\usepackage[acronym]{glossaries}
\usepackage{subfigure}
\usepackage{tabularx}
\usepackage{threeparttable}
\usepackage{nicefrac}
\usepackage{microtype}

\newcommand{\qs}{\Delta}
\newcommand{\qp}{q}
\newcommand{\X}{\mathbf{X}}
\newcommand{\Y}{\mathbf{Y}}
\newcommand{\splreg}{\mathcal{R}}

\newcommand{\fc}{\mathbf{f}_\text{C}}
\newcommand{\fq}{\mathbf{f}_\text{Q}}
\newcommand{\fcq}{\mathbf{f}}
\newcommand{\Fcq}{\mathbf{F}}
\newcommand{\D}{\mathcal{D}}
\newcommand{\Hcq}{\mathbf{H}}
\newcommand{\barHcq}{\bar{\mathbf{H}}}
\newcommand{\ver}{\text{VER}}
\newcommand{\Dtemp}{\D_\text{temp}}
\newcommand{\Dspat}{\D_\text{spat}}
\newcommand{\cpos}{c}
\newcommand{\jump}{j}
\newcommand{\kersize}{w}
\newcommand{\nfeat}{m}
\newcommand{\pad}{z}
\newcommand{\recfield}{r}
\newcommand{\stride}{s}

\newtheorem{definition}{Definition}
\theoremstyle{definition}

\setacronymstyle{long-short}
\newacronym{ai}{AI}{artificial intelligence}
\newacronym{auc}{AUC}{area under the curve}
\newacronym{bn}{BN}{batch normalization}
\newacronym{cnn}{CNN}{convolutional neural network} 
\newacronym{dct}{DCT}{discrete cosine transform}
\newacronym{focal}{FOCAL}{FOrgery loCALizer}
\newacronym[longplural={groups-of-pictures}]{gop}{GOP}{group-of-pictures}
\newacronym{hmrf}{HMRF}{Huber Markov random field}
\newacronym{hog}{HoG}{histogram of oriented gradients}
\newacronym{lstm}{LSTM}{long short-term memory} 
\newacronym{macemrh}{MACE-MRH}{minimum average correlation energy Mellin radial harmonic}
\newacronym{pr}{PR}{precision-recall}
\newacronym{relu}{ReLU}{rectified linear unit}
\newacronym{rewind}{REWIND}{REVerse engineering of audio-VIsual coNtent Data}
\newacronym{rnn}{RNN}{recurrent neural network}
\newacronym{roc}{ROC}{receiver operating characteristic}
\newacronym{sgdm}{SGDM}{stochastic gradient descent with momentum}
\newacronym{sift}{SIFT}{scale-invariant feature transform}
\newacronym{sulfa}{SULFA}{Surrey University Library for Forensic Analysis}
\newacronym{ver}{VER}{variance-to-entropy ratio}

\journal{}

\begin{document}

\begin{frontmatter}



\title{FOCAL: A Forgery Localization Framework based on Video Coding Self-Consistency}

\tnotetext[]{This work has been partially supported by the University of Padova project Phylo4n6 prot. BIRD165882/16. This material is based on research sponsored by DARPA and Air Force Research Laboratory (AFRL) under agreement number FA8750-16-2-0173. The U.S. Government is authorized to reproduce and distribute reprints for Governmental purposes notwithstanding any copyright notation thereon. The views and conclusions contained herein are those of the authors and should not be interpreted as necessarily representing the official policies or endorsements, either expressed or implied, of DARPA and Air Force Research Laboratory (AFRL) or the U.S. Government.}

\author[unipd]{Sebastiano Verde}
\author[polimi]{Paolo Bestagini}
\author[unipd]{Simone Milani}
\author[unipd]{Giancarlo Calvagno}
\author[polimi]{Stefano Tubaro}

\address[unipd]{Department of Information Engineering, University of Padova, Padua, Italy}
\address[polimi]{Dipartimento di Elettronica, Informazione e Bioingegneria, Politecnico di Milano, Milan, Italy}

\begin{abstract}
	\input{abstract.tex}
\end{abstract}

\begin{keyword}
Forgery Detection \sep
Video Splicing \sep 
Content Integrity \sep
Feature Fusion \sep
Multimedia Forensics



\end{keyword}

\end{frontmatter}

\input{intro}
\input{related}
\input{problem}
\input{network}
\input{algo}
\input{results}

\input{conclusion}

{
\bibliographystyle{elsarticle-num} 
\bibliography{biblio}
}

\end{document}

%% file: abstract.tex
Forgery operations on video contents are nowadays within the reach of anyone, thanks to the availability of powerful and user-friendly editing software. Integrity verification and authentication of videos represent a major interest in both journalism (e.g., fake news debunking) and legal environments dealing with digital evidence (e.g., a court of law). While several strategies and different forensics traces have been proposed in recent years, latest solutions aim at increasing the accuracy by combining multiple detectors and features.
This paper presents a video forgery localization framework that verifies the self-consistency of coding traces between and within video frames, by fusing the information derived from a set of independent feature descriptors. The feature extraction step is carried out by means of an explainable \acrlong{cnn} architecture, specifically designed to look for and classify coding artifacts. The overall framework was validated in two typical forgery scenarios: temporal and spatial splicing. Experimental results show an improvement to the state-of-the-art on temporal splicing localization and also promising performance in the newly tackled case of spatial splicing, on both synthetic and real-world videos.

%% file: intro.tex
\section{Introduction}\label{sec:intro}
Authenticity assessment for video sequences is nowadays a paramount task in several contexts, such as citizen journalism and fake news debunking, as well as evidence validation in legal procedures and fraud detection. This concern has gained importance during the last years because of the wide availability of powerful and easily-operable video editing programs (e.g., Adobe Premiere, Apple Final Cut, etc.) and the wide-spread use of video data in communication and documenting activities. Moreover, the development of deep learning solutions for the automatic creation and editing of image and video contents have posed new challenges to forensic analysts, since a malicious user has the opportunity to create fake contents that overcome most of the existing detectors.

As a matter of fact, forensic analysts have been constantly investigating innovative and accurate solutions for forgery detection and localization. Among the first strategies being proposed, we can find detectors that identify the acquisition device \cite{Chen2007, Bayram2008}, physical inconsistencies \cite{Conotter2012}, video recapturing \cite{VisentiniScarzanella2012}, frame deletion and insertion \cite{Stamm2012, Bestagini2013}, or codec-related operations \cite{Bian2014, Bestagini2016, Milani2012}. Most of these detectors verify the \emph{self-consistency} \cite{Huh18:fakenews_selfconsist, mat16:video_forgery_corr} of video processing footprints, i.e., the uniformity of traces left on the signal across different frames and regions of the video sequence. Whenever an external element is included within an original image or video, the forensic footprints in the altered region change with respect to untouched ones. Revealing such a discrepancy allows detecting the possible presence of a forgery. 

\begin{figure*}
\includegraphics[width=\textwidth]{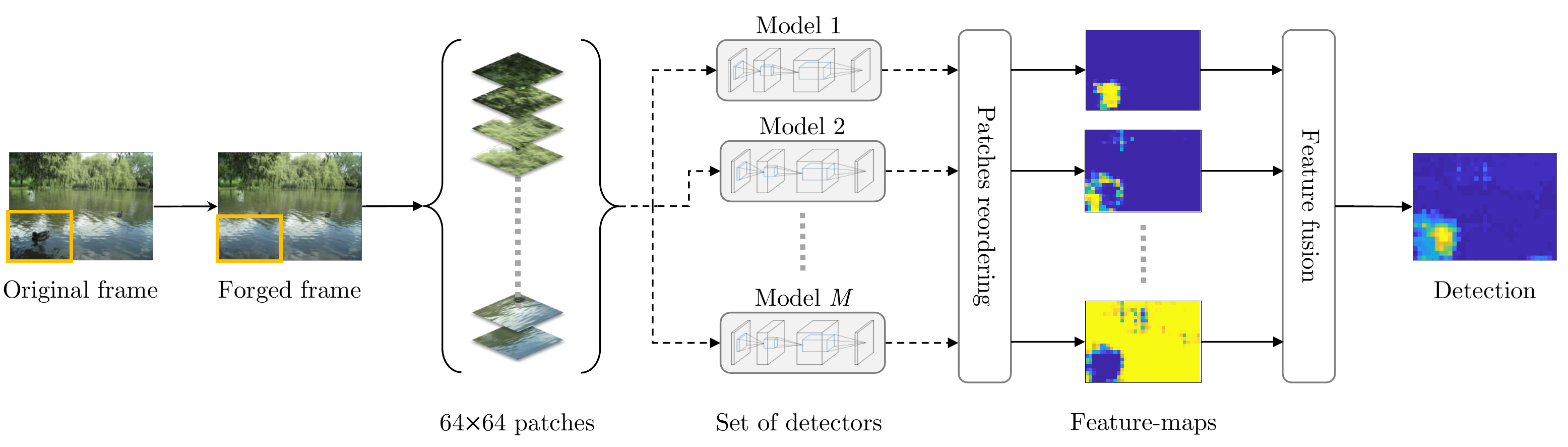}
\caption{\Acrfull{focal} framework. A forged video-frame is split into 64$\times$64 patches and fed to a set of pre-trained detectors (e.g. classifiers of the video coding standard and quality). Extracted features are rearranged into feature-maps and a fusion function merges them into a single detection heat-map. Dashed and solid lines are used to denote patch-wise and frame-wise operations, respectively.}
\label{fig:scheme}
\end{figure*}

Extending the preliminary work in \cite{Ver18:video_splicing}, the current paper proposes a \gls{focal} that checks the self-consistency of multiple and independent forensic traces related to video coding (Figure \ref{fig:scheme}). Differently from the previous work, where forgeries only consisted in concatenating video sequences from different sources (\emph{temporal splicing}), this new approach is also able to precisely localize an altered region within a single frame (\emph{spatial splicing}) as well as along time dimension.

Given an input video, each frame is split into smaller patches, and a feature vector is extracted from each one of them. The set of features corresponds to the output values of the final softmax layers from multiple \glspl{cnn} dedicated to the classification of different coding parameters, such as coding standard and quality level. These \glspl{cnn} share an \emph{explainable} architecture, which was specifically designed to look for coding artifacts by aligning the receptive fields of the network filters to the quantization block boundaries, where the most significant traces are typically visible.

An unsupervised fusion technique was designed to merge the outputs of these heterogeneous feature descriptors into a human-readable heatmap, which characterizes each frame-patch from the analyzed video with a likelihood measure that models the probability of being forged. This approach also makes the framework scalable and extendible at will, allowing the introduction of additional detectors and feature descriptors to contribute to the overall heatmap. 

Experimental validation takes into account different forgery setups, such as temporal and spatial splicing, in controlled and uncontrolled environments. Results show that the proposed solution is able to improve the performance of \cite{Ver18:video_splicing}, thanks to the newly adopted network architecture, and to obtain convincing results in the detection of local forgeries as well, with an \gls{auc} of the \gls{roc} curve of $0.94$.

The rest of the paper is organized as follows. Section~\ref{sec:related} briefly overviews the literature on video forgery detection and localization, distinguishing between temporal and spatial forgeries. Section \ref{sec:problem} formally defines the problem addressed by the paper and the notation used. Section~\ref{sec:net} presents the proposed \gls{cnn} for extracting coding-related features, with special emphasis on the architectural choices. Section~\ref{sec:algo} illustrates our forgery localization framework, from the feature extraction step to the final feature fusion and heatmap generation, in both temporal and spatial forgery cases. Section~\ref{sec:results} reports all the details about the experimental setup, the training phase, the generation of the synthetic dataset and the obtained results. Finally, Section~\ref{sec:conclusion} concludes the paper and outlines possible future work.

%% file: related.tex

\section{Related work}\label{sec:related}



In recent years, video authentication has emerged as a novel and challenging research field \cite{Milani2012a, Sit16:video_forgery_passive}  leading to the development of algorithms and tools capable of estimating whether a video sequence is original or not. Most of the proposed approaches identifies two different types of forgeries: temporal and spatial splicing. The first case affects a video through the inclusion or deletion of some frames into or from the original sequence. In the second one, the content of individual frames is modified, e.g., with a cut-and-paste alteration of a region (inclusion/removal of an object from the scene) or performing an upscale-crop editing (where an object located in an outermost part of the video is removed by cropping the frames). Furthermore, it is worth mentioning double/multiple compression. Since video sequences are usually available in compressed format and their alterations are carried out in the pixel domain, videos must be re-encoded every time a forgery is operated. For this reason, forged sequences typically exhibit the presence of multiple compression artifacts. 

Among the strategies for detecting the insertion/deletion of frames, some algorithms  exploit the correlation and similarity between frame characteristics  \cite{Yang2016, Lin11:video_tamp_color}: if some temporal patterns do not follow the expected trend, the algorithm raises an alarm. Similarly, other solutions identify regular patterns in the camera noise signals: whenever there are repetitions \cite{De07:video_forgery} or oddities  \cite{Mon07:video_forg_noise, Kob10:video_forgery_noise} due to the fact that forged frames were taken by a different camera, the algorithm reports an anomaly. Deletions can be highlighted by spotting irregularities in motion statistics, obtainable through the analysis of the optical flow \cite{Wan14:video_forged_oflow, Kingra2017}, interpolation \cite{Bes13:temp_interp}, or standard block matching \cite{Su09:video_forgery_mcea}. The correlation in prediction residual information \cite{Bes13:video_tamp_res, Liu14:frame_del}, in texture patterns \cite{Zha15:video_tamp_lbp} and in brightness \cite{Zhe15:video_forged_bright} can be exploited as well.

Spatial forgeries can be detected by checking the consistency of forensic traces left by the acquisition device or different encoding algorithms on the video sequence. The strategy proposed in \cite{Hiu13:upscalecrop_detection} exploits the scaling invariance of the \gls{macemrh} correlation filter to unveil traces of upscale-crop forgeries. Similarly, the impact of a spatial splicing on interlaced videos can be analyzed to reveal the traces of a possible alteration \cite{Wan07:video_foreged_interlaced}. 
Object removal is exposed by detecting discrepancies in the motion vectors \cite{Li13:detect_removed_object} and through a combination of different steganalysis features \cite{Che16:object_removal_ensemble}.
Copy-move object removal can be revealed by exploiting local descriptors, such as \gls{hog} \cite{Sub12:copymove_hog} or \gls{sift} descriptors \cite{Ame10:sift_copy_move}. The same type of forgery can also be detected by analyzing the spatial and temporal correlation among frames \cite{Pan14:copy_move_spatiotemp}, Zernike moments \cite{Dam15:copy_move_zernicke}, and optical flow similarities \cite{Bid15:copy_move_oflow}.
Together with intra-frame similarities and discrepancies, it is possible to expose physical inconsistencies in scene illumination and object motion, by comparing a plausible model with what is estimated directly from the pixels \cite{Conotter2012}.

Revealing traces of multiple compression on the analyzed video sequence allows an effective detection of editing operations. A first insight was provided in \cite{Wan06:video_forgery_mpeg}, followed by several researches extending to videos the coding footprints identified for images \cite{Milani2012, Xu13:double_mpeg_detection}. The misalignment of the \gls{gop} structures related to the first and the second encoding can be informative as well: whenever the coding parameters change between two consecutive encoding steps, a superposition of heterogeneous artifacts appears on the video and can be detected \cite{Vaz12:video_forged_gop}. Similarly, the simultaneous presence of traces related to incompatible coding parameters or formats is investigated in several papers \cite{Bestagini2016, Bestagini2012, Hua14:double_video_comp}. Furthermore, whenever a video sequence is compressed twice, it is possible to observe some peculiar noise patterns: in \cite{Jia13:double_mpeg_noise} the authors propose a first-order Markov statistics for the differences between quantized \gls{dct} coefficients along different directions, while the solution in \cite{Rav14:video_forgery_noise} employs a modified \gls{hmrf} model. These methods enables to assess whether a whole sequence of frames is authentic or tampered (e.g. compressed twice) but, differently from the proposed one, do not allow for precisely localizing a forgery operation.

Recent approaches are setting up a new trend in video forgery detection by using deep neural networks. In \cite{d2017autoencoder} the authors propose an autoencoder structure to learn a synthetic model of the source (forgeries are detected as outlier of the learned model) followed by \glspl{rnn}, implemented with the \gls{lstm} model, to exploit temporal dependencies. In \cite{Ver18:video_splicing} two \glspl{cnn} are independently trained to extract codec- and quality-related features with the purpose of detecting temporal inconsistencies, showing that the combination of heterogeneous detectors enhances the overall performance. Some studies have also been addressed to expose the newly appeared threat of \acrshort{ai}-generated highly-realistic forgeries, also known as DeepFakes. The detection is carried out by means of eye-blinking analysis \cite{li2018ictu} and combinations of deep learning models such as \glspl{cnn} and \glspl{rnn} \cite{guera2018deepfake}.    

Following the trend of fusing multiple features to increase the overall accuracy, the proposed strategy combines a set of coding-related features obtained from different \glspl{cnn}. The architecture of these networks was designed in awareness of where and how compression artifacts appear, as described in Section \ref{sec:net}.

%% file: problem.tex
\section{Problem definition}\label{sec:problem}

The purpose of \gls{focal} framework is detecting and localizing temporal and spatial splicing operations on video sequences. Here we provide a formal definition of the tackled problem and the notation used throughout the paper. 

Let us define a video sequence $\X$ as an array of $N$ frames denoted by $\X_n$,
\begin{equation}
\X = \left[ \X_1, \X_2, \ldots, \X_N\right],
\end{equation}
where each frame is a matrix of pixels $X_{uv}$,
\begin{equation}
\X_n = \left[ X_{uv} \right]_n. 
\end{equation}
Pixel coordinates are $(u,v)\in\mathcal{U}\times\mathcal{V}$, where $U=|\mathcal{U}|$ and $V=|\mathcal{V}|$ are the frame dimensions.

\begin{definition}
Let $\X$ and $\Y$ be two sequences of frames of dimensions $U_X$, $V_X$, $N_X$ and $U_Y$, $V_Y$, $N_Y$, respectively. The two sequences are \emph{spliceable} if $U_X=U_Y$ and $V_X=V_Y$.
\end{definition}

Without loss of generality, we will define the two types of forgeries addressed by this work for spliceable sequences only.

\begin{definition}
Let $\X$ and $\Y$ be two spliceable sequences. A \emph{temporal splicing} is a function $\mathcal{T}$ that concatenates $\X$ and $\Y$ into a single sequence:
\begin{equation}\label{eq:temp_splice}
\mathcal{T}\left(\X,\Y\right) = \left[\X_1,\ldots,\X_{N_X},\Y_1,\ldots,\Y_{N_Y}\right].
\end{equation}
The resulting sequence is called \emph{temporally-spliced} and the frame-index $N_X+1$ is the \emph{splicing point}.
\end{definition}

\begin{definition}
Let $[X_{uv}]$ and $[Y_{uv}]$ be two frames from two spliceable sequences $\X$ and $\Y$, with pixels in $\mathcal{U}\times\mathcal{V}$. Let $\splreg\subset\mathcal{U}\times\mathcal{V}$ be a subset of pixel coordinates.  A \emph{spatial splicing} is a function $\mathcal{S}$ that substitutes pixels in $\splreg$ of one frame with pixels in $\splreg$ of the other frame:
\begin{equation}
\mathcal{S}\left(X_{uv},Y_{uv},\splreg\right) =
\begin{cases}
Y_{uv},\quad (u,v)\in\splreg\\
X_{uv},\quad \mathrm{otherwise}
\end{cases}.
\end{equation}
The resulting sequence is called \emph{spatially-spliced} and the altered region $\splreg$ is called \emph{spliced region}.
\end{definition}

Given a video sequence under analysis, with no additional information available except for the pixel values, we aim to localize possible splicing points or spliced regions.

%% file: network.tex
\section{Coding features}\label{sec:net}

\begin{figure}
\centering
	\subfigure[][High quality.]{
		\includegraphics[width=0.473\columnwidth]{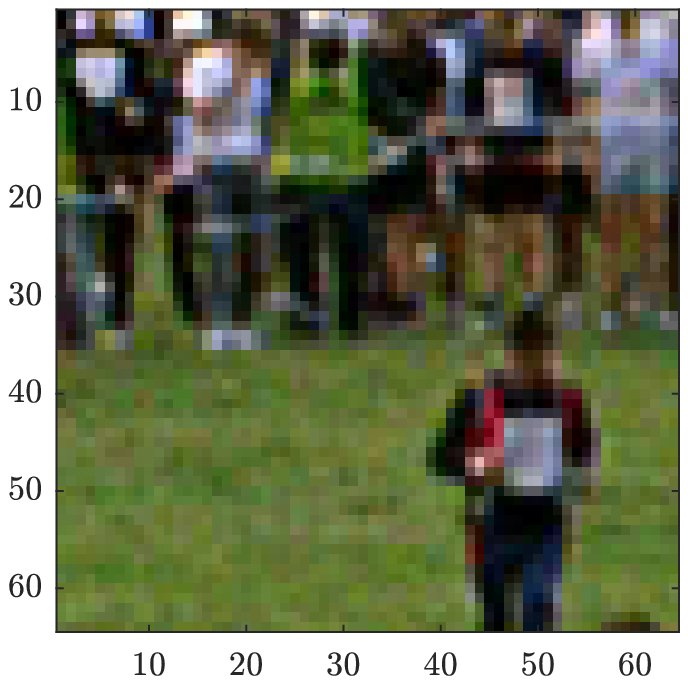}
		\label{fig:patch_hq}}%
	\subfigure[][Low quality.]{
		\includegraphics[width=0.473\columnwidth]{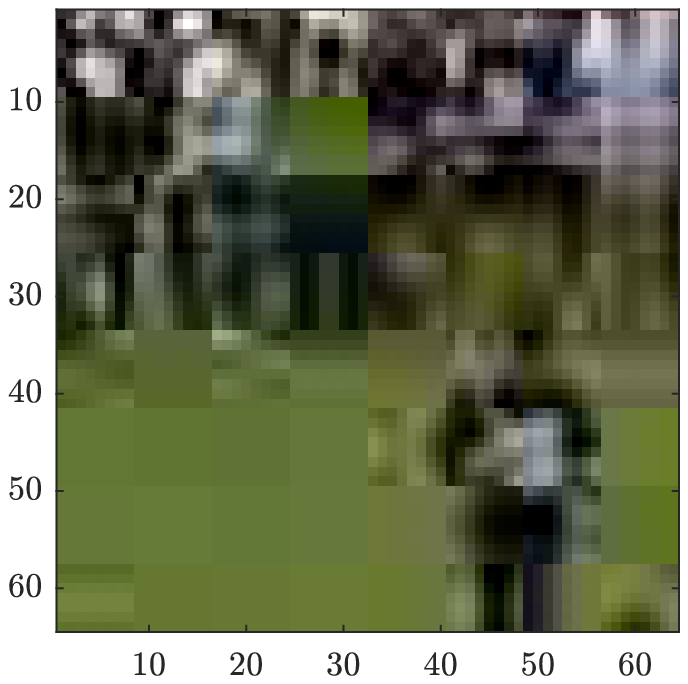}
		\label{fig:patch_lq}}%
\caption{Block-artifacts at different encoding qualities in a $64\times64$ patch. The grid of $8\times 8$ blocks is particularly evident in \ref{fig:patch_lq}.}
\label{fig:blocks}
\end{figure}

The core of our forgery localizer consists of a \acrlong{cnn} specifically designed to detect and classify traces left by video-coding algorithms. Understanding which coding scheme and parameters were used to encode a video clip, by only looking at the pixel domain, represents a challenging task even for a human observer. However, almost anyone is able to perform a rough classification on the perceptive quality of a video, usually by looking for the presence of block-artifacts, typically more evident in lower quality videos (see Figure \ref{fig:blocks}).

Block-artifacts are introduced by any coding algorithm adopting the block-based transform principle. This coding paradigm first splits the set of frames into \acrfullpl{gop} that are encoded independently of one another. Each frame within a \gls{gop} is encoded according to a pattern of predefined types.
\begin{itemize}
\item \emph{Intra} or I type: the frame is coded independently from all the others; the first frame of each \gls{gop} must be intra-coded.
\item \emph{Predictive} or P type: the frame is encoded with motion compensation, using the previous I or P frames as references.
\item \emph{Bidirectionally predictive} or B type: the frame is encoded with motion compensation, using the previous and the following I or P frames as references.
\end{itemize}  

The actual encoding procedure involves a domain transform (for example the \gls{dct}) applied to block of pixels, typically $8\times8$. The obtained coefficients are then quantized and fed into an entropy encoder. The lower the bitrate, the coarser is the coefficients representation, resulting in blurry blocks and evident discontinuities at the block boundaries. 

Since block-artifacts appear to be quite distinctive for the human vision, we attempted to design a network whose attention is \emph{focalized} on these particular features. Specifically, we wanted our network to analyze the regions nearby the \emph{corners} of the block-grid, as each one of them allows to observe four blocks and boundaries at the same time. To accomplish that, the network architecture was designed to align with the block-grid and to extract a descriptor for each corner and its associated neighborhood. To better understand this, we need to introduce the concept of \emph{receptive field}.
%
%
\newcommand{\laysz}{0.31}
\begin{figure*}[t]
\centering

	\subfigure[][$\nfeat_0=64$; $\jump_0=1$; $\recfield_0=1$; $\cpos_0=0.5$]{
		\includegraphics[width=\laysz\textwidth]{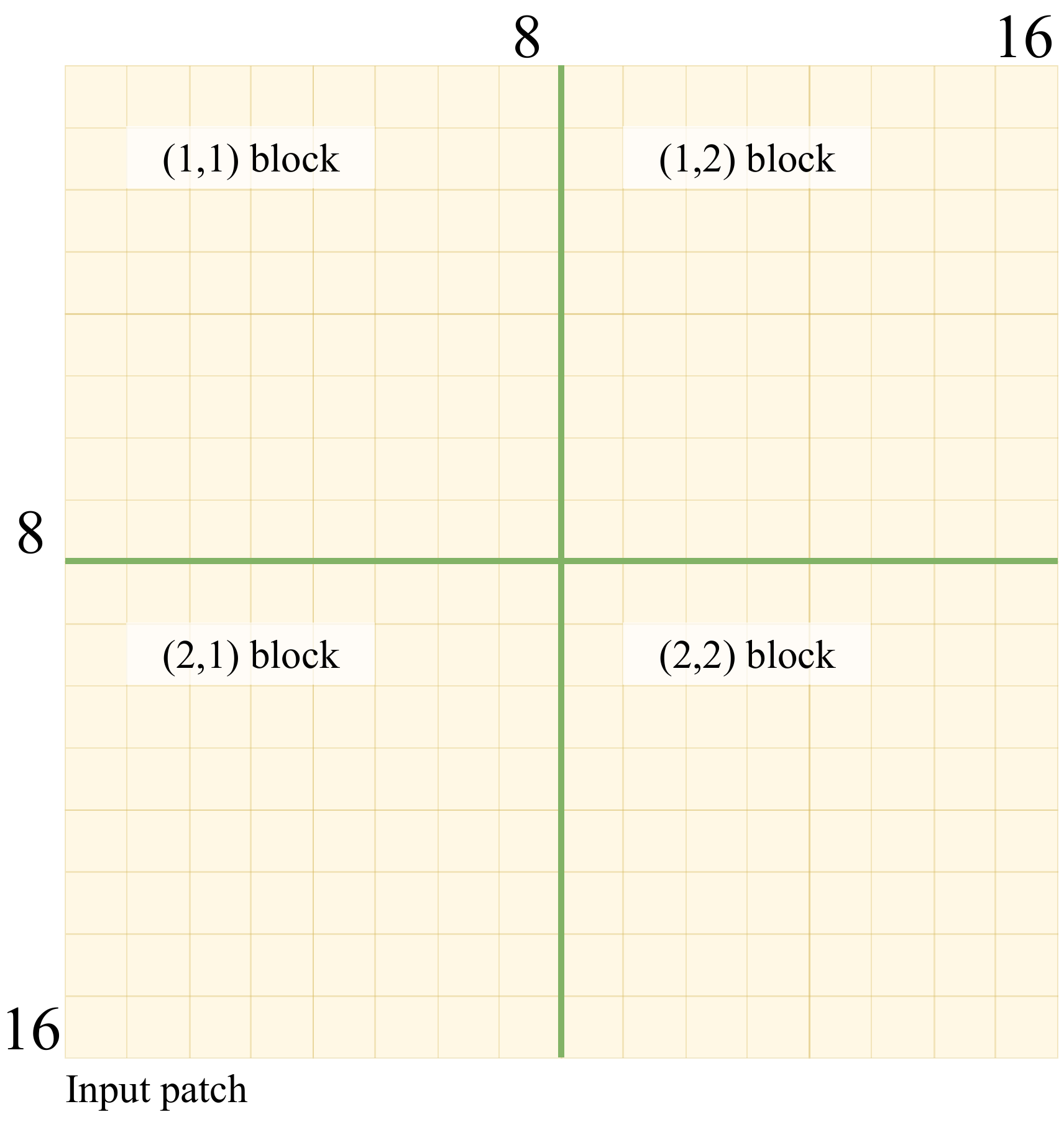}
		\label{fig:inlayer}}%
	\subfigure[][$\nfeat_1=61$; $\jump_1=1$; $\recfield_1=4$; $\cpos_1=2$]{
		\includegraphics[width=\laysz\textwidth]{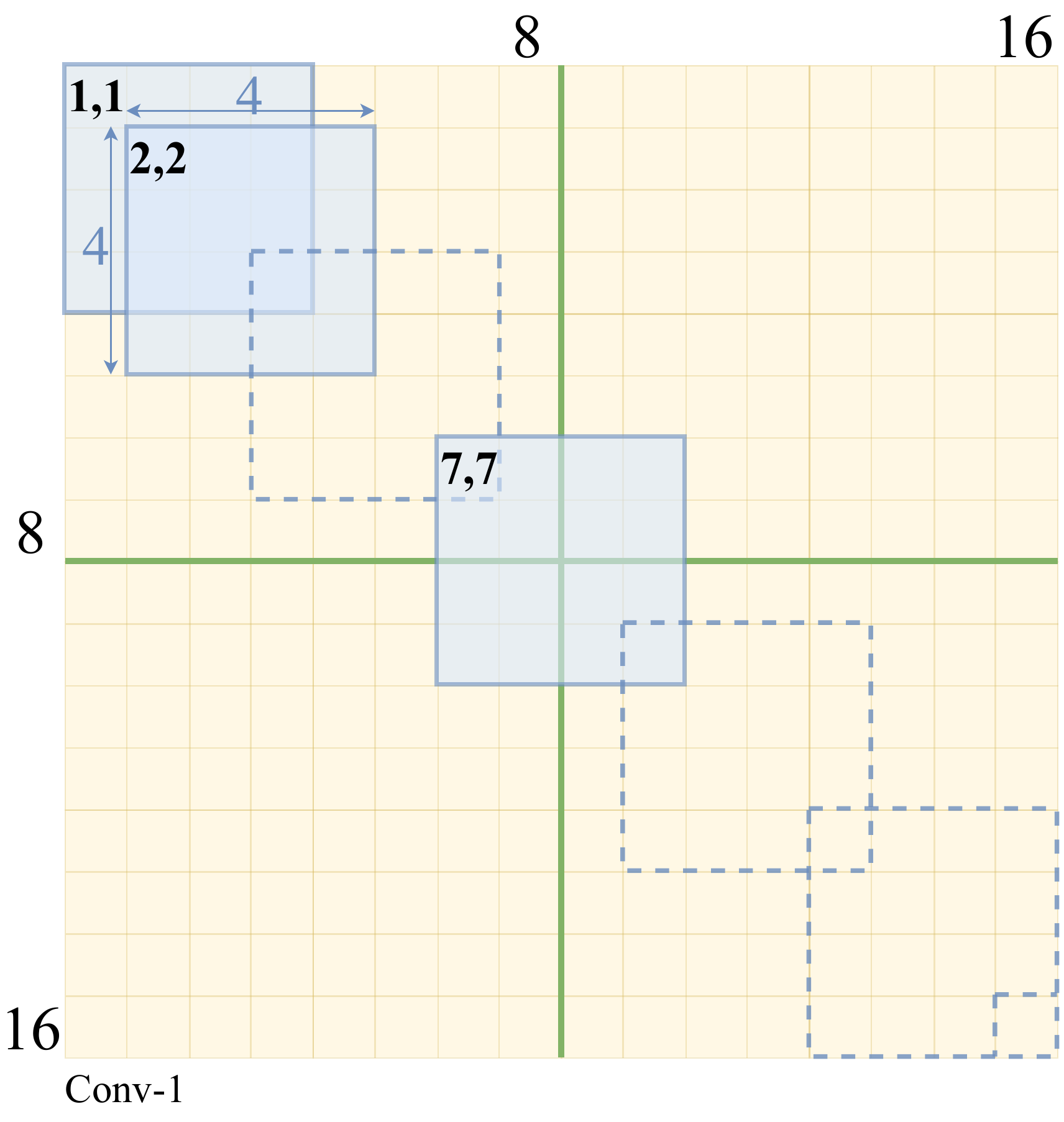}
		\label{fig:conv1}}%
	\subfigure[][$\nfeat_2=30$; $\jump_2=2$; $\recfield_2=6$; $\cpos_2=3$]{
		\includegraphics[width=\laysz\textwidth]{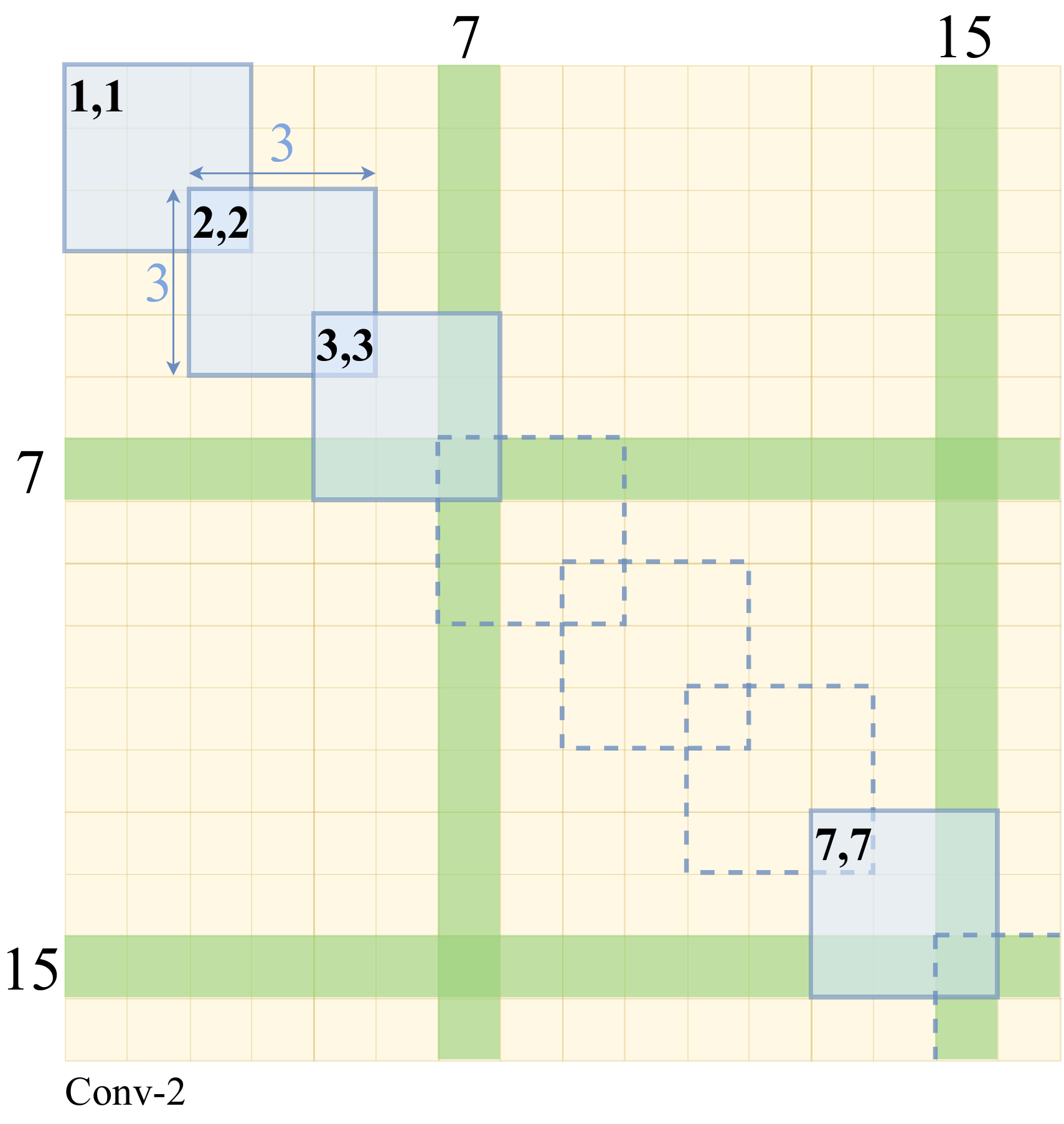}
		\label{fig:conv2}}\hfill
		
	\subfigure[][$\nfeat_3=27$; $\jump_3=2$; $\recfield_3=12$; $\cpos_4=6$]{
		\includegraphics[width=\laysz\textwidth]{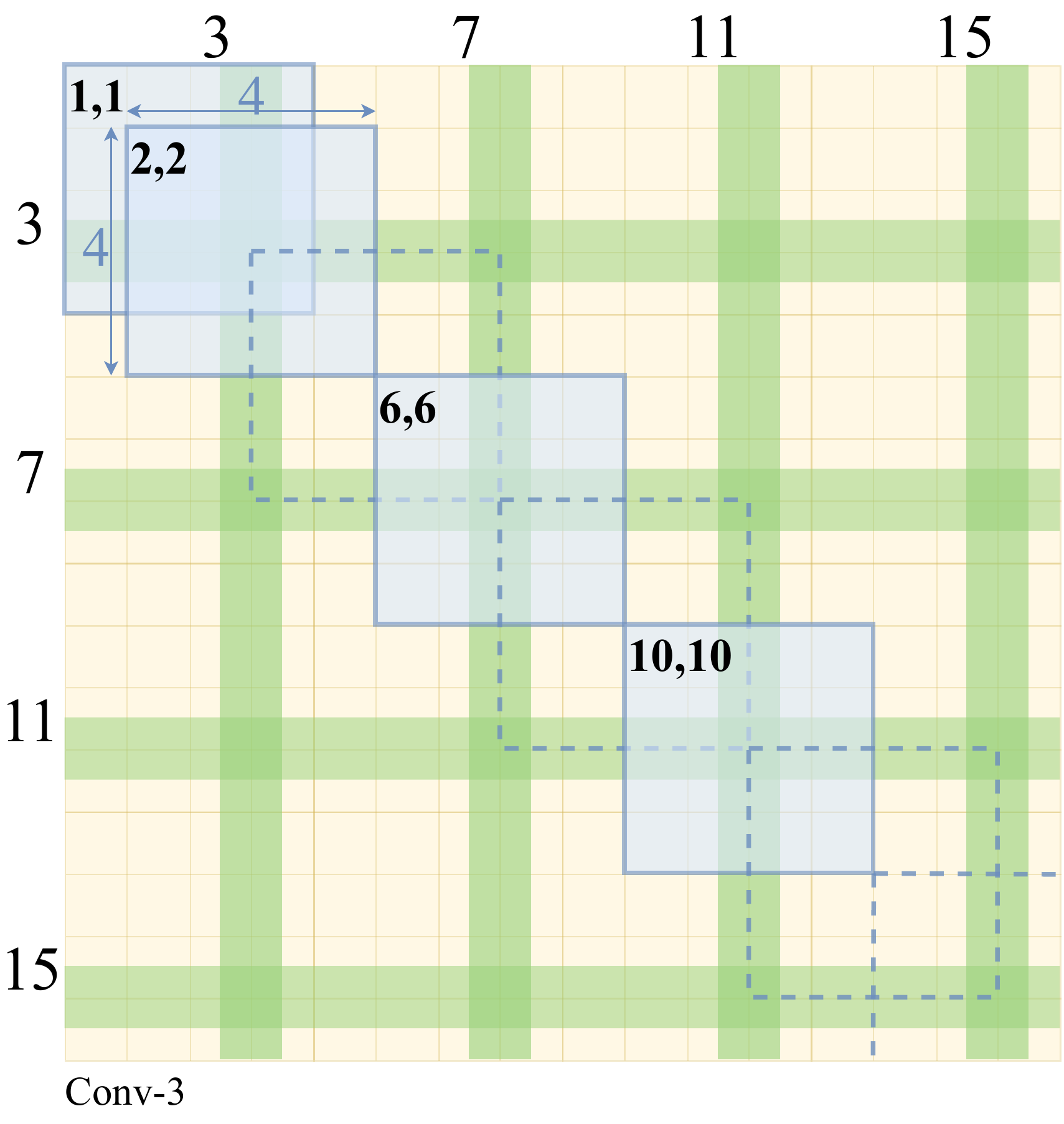}
		\label{fig:conv3}}%
	\subfigure[][$\nfeat_4=13$; $\jump_4=4$; $\recfield_4=16$; $\cpos_4=8$]{
		\includegraphics[width=\laysz\textwidth]{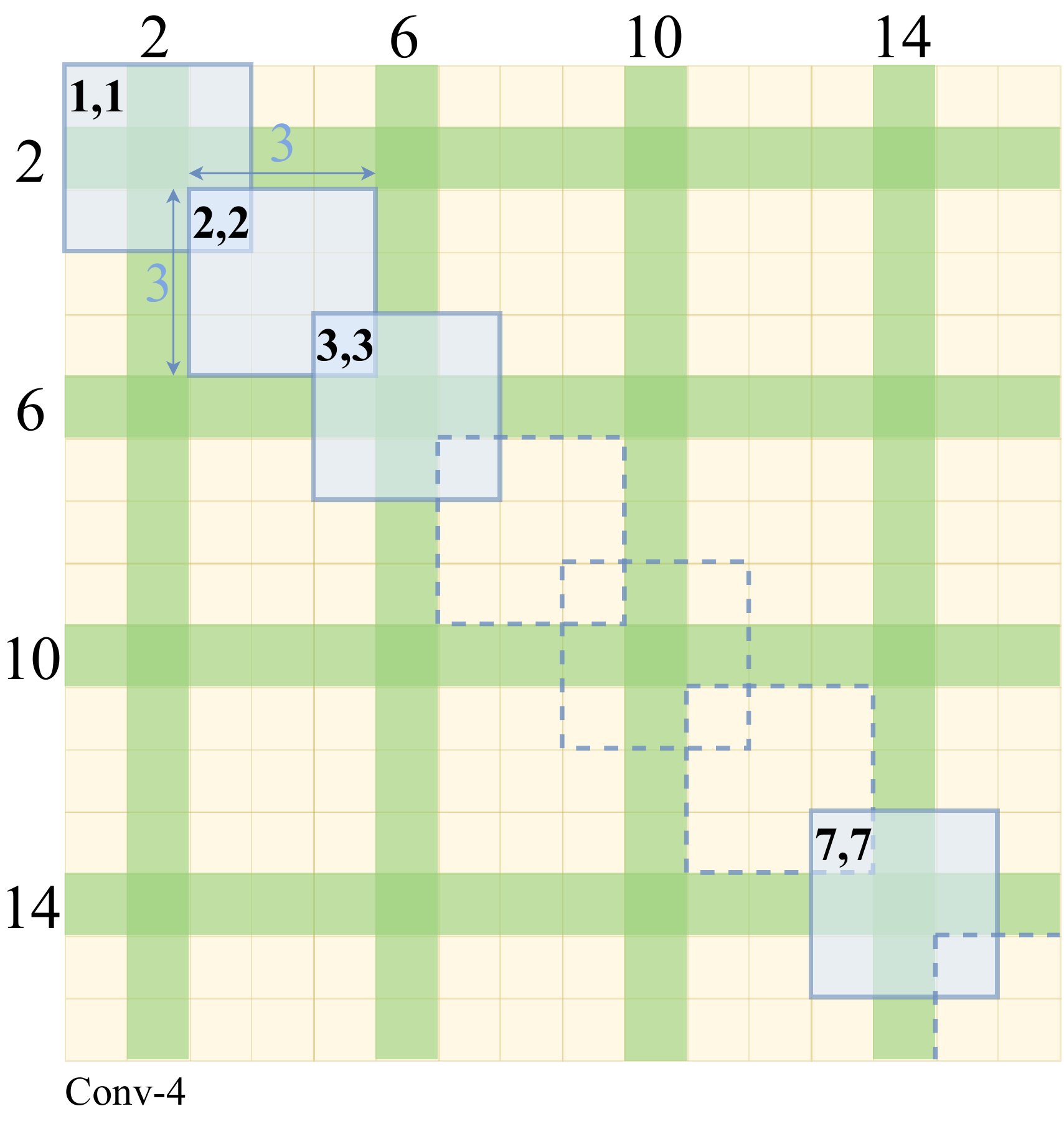}
		\label{fig:conv4}}%
	\subfigure[][$\nfeat_5=7$; $\jump_5=8$; $\recfield_5=24$; $\cpos_5=8$]{
		\includegraphics[width=\laysz\textwidth]{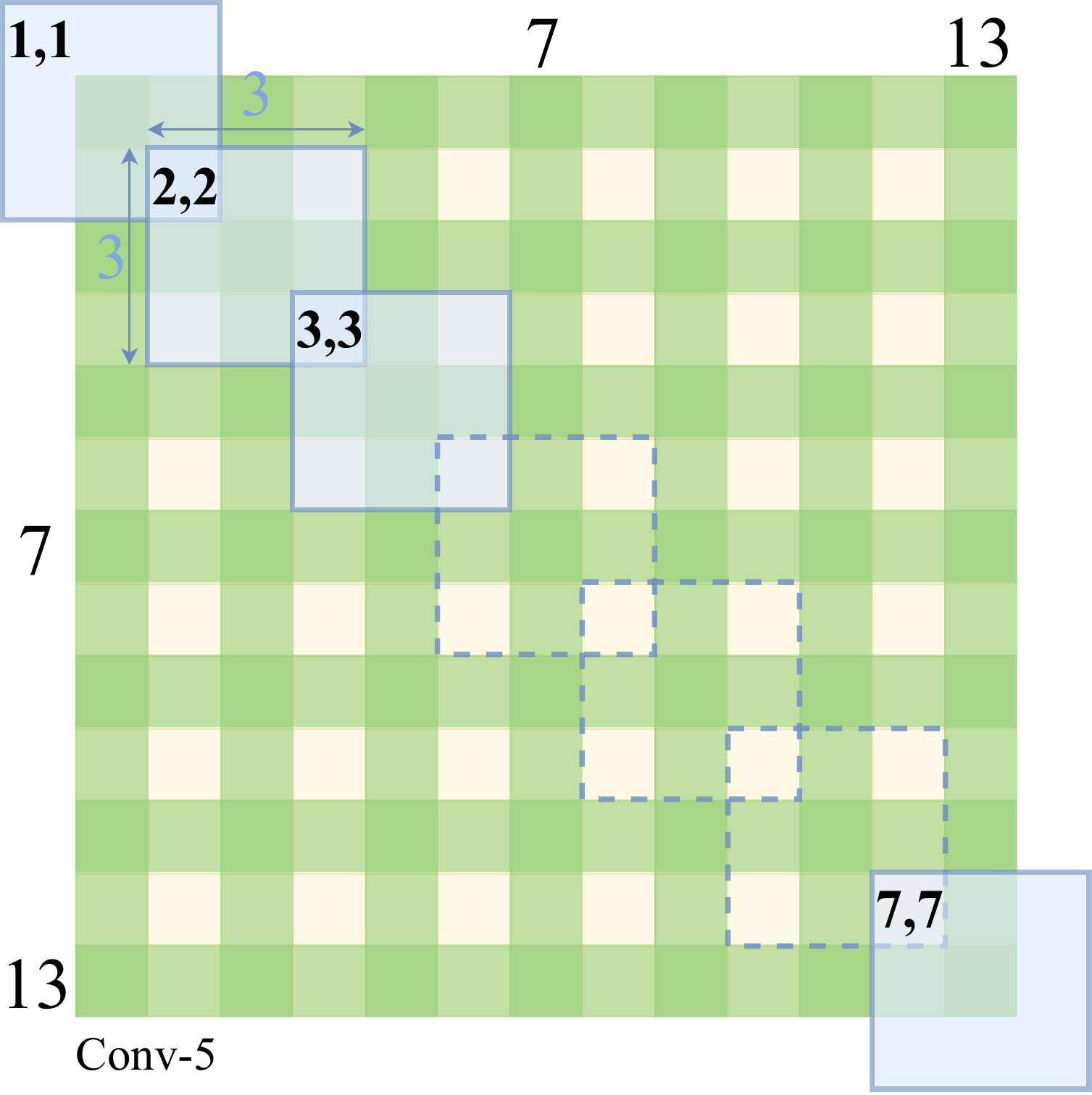}
		\label{fig:conv5}}
		
\caption{Architecture of the convolutional layers, with feature map size ($\nfeat_i$), jump factor ($\jump_i$), receptive field size ($\recfield_i$) and center position ($\cpos_i$). Green areas denote feature activations related to block boundaries. The output of Conv-5 is a 7-by-7-by-64 tensor, consisting of one 64-length feature vector for each corner of the block grid.}
\label{fig:layers}
\end{figure*}
The receptive field denotes the region of the input that a particular network neuron is looking at. It is described by its center position and its size. Pixels contribution to the calculation of the output feature grows exponentially towards the center of the receptive field. Given the input size, the layers of a \gls{cnn} can be designed in order to produce features with the desired receptive field. 

The \gls{cnn} architecture we propose consists of: five convolutional layers, each one followed by a \gls{bn} and a \gls{relu} activation; two fully-connected layers; a softmax activation layer. Table \ref{tab:net_arch} reports the complete list of layers, specifying for each convolutional one the number of kernels, the kernel size $\kersize$, the stride $\stride$ and the padding size $\pad$.

\begin{table}
\centering
\begin{threeparttable}
\caption{Network architecture}\label{tab:net_arch}
\begin{tabular}{lccccc}
\toprule
Layer & Kernels & $\kersize$ & ~$\stride$~ & ~$\pad$~ & Activation\\
\midrule
Conv-1~ & 64 & ~$4\times4$~ & 1 & 0 & ~\gls{bn} + \gls{relu}~\\ 
Conv-2~ & 64 & $3\times3$ & 2 & 0 & ~\gls{bn} + \gls{relu}~\\ 
Conv-3~ & 64 & $4\times4$ & 1 & 0 & ~\gls{bn} + \gls{relu}~\\ 
Conv-4~ & 64 & $3\times3$ & 2 & 0 & ~\gls{bn} + \gls{relu}~\\ 
Conv-5~ & 64 & $3\times3$ & 2 & 1 & ~\gls{bn} + \gls{relu}~\\ 
\midrule
FC-1 & 64 & & & &\\
FC-2 & $K$ & & & & Softmax\\
\bottomrule
\end{tabular}
\begin{tablenotes}
\item $\kersize$ = kernel size; $\stride$ = stride; $\pad$ = padding.
\end{tablenotes}
\end{threeparttable}
\end{table}

The input to the \gls{cnn} is a luminance patch of $64\times64$ pixels. Chrominances are neglected since they do not add relevant information to block-artifacts and are often subsampled. Assuming a block-grid of $8\times8$ transform blocks, each patch contains exactly $7\times7=49$ corners.

Figure \ref{fig:layers} provides a visualization of the five convolutional layers and a computation of four geometrical parameters per each layer:
\begin{itemize}
\item the number of output features $\nfeat$, based on the number of input features and the layer properties,
\begin{equation}
\nfeat_{\textit{out}} = \left\lfloor \frac{\nfeat_{\textit{in}}+2\pad-\kersize}{\stride}\right\rfloor+1;
\end{equation}
\item the jump factor $\jump$ in the output feature map, given by the jump in the input times the stride size,
\begin{equation}
\jump_{\textit{out}} = \jump_{\textit{in}} \cdot \stride;
\end{equation}
\item the receptive field size $\recfield$ of the output feature map,
\begin{equation}
\recfield_{\textit{out}} = \recfield_{\textit{in}} + (\kersize-1)\cdot \jump_{\textit{in}};
\end{equation}
\item the center position $\cpos$ of the receptive field of the first output feature,
\begin{equation}
\cpos_{\textit{out}} = \cpos_{\textit{in}} + \left(\frac{\kersize-1}{2}-\pad\right)\cdot \jump_{\textit{in}}.
\end{equation}
\end{itemize}
All parameters are computed with respect to those of the previous layer. In the input layer, we have $\nfeat_0=64$ features (the input size), jump factor and receptive field both equal to one pixel ($\jump_0=\recfield_0=1$) and the center position is the center of the first pixel ($\cpos_0=0.5$).
Green areas in Figure \ref{fig:layers} represent those parts of the feature map carrying information related to block boundaries. Yellow areas are associated with pixels within the blocks. 

With this particular design, the network progressively condenses the block-grid, without blending together the descriptors associated to different corner points. The output of the last convolutional layer is a 7-by-7-by-64 tensor, forming a map of 64-elements descriptors, one for each corner of the input patch. This tensor is fed into a fully connected network that returns the final $K$-length patch descriptor, where $K$ depends on the chosen number of classification outputs. 

The feature descriptors calculated by this \gls{cnn} can be exploited in a variety of forensic applications. In the following section, we discuss the design of our forgery localization framework, which checks the self-consistency of these coding features to detect temporal and spatial splicing operations.

%% file: algo.tex

\section{Forgery localization}\label{sec:algo}

\begin{figure*}
\centering

	\subfigure[t][Medium-low-quality MPEG-2 spliced with low-quality MPEG-4.]{
		\includegraphics[width=0.486\textwidth]{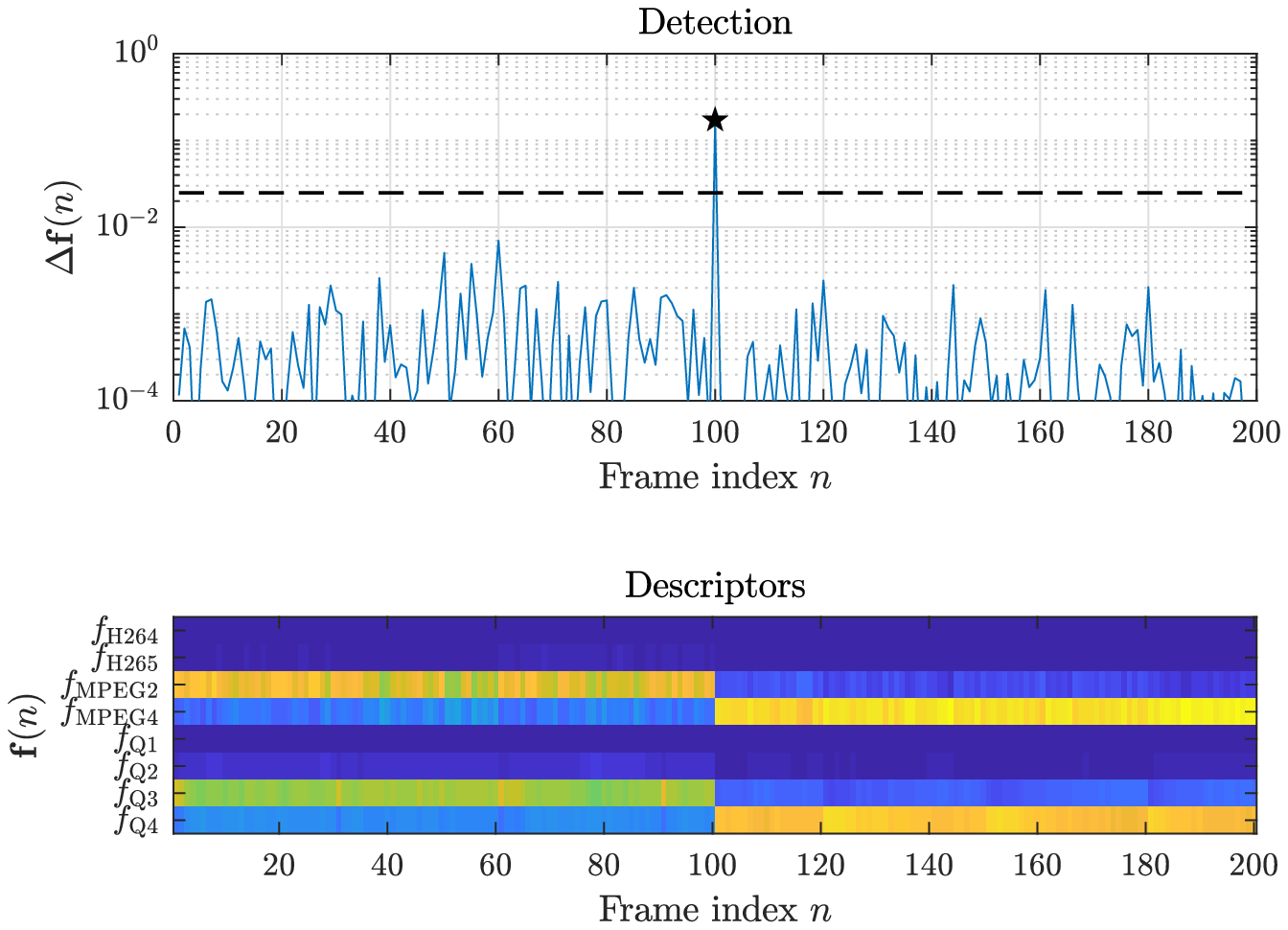}
		\label{fig:splicing_ex}}%
	\subfigure[t][High-quality H.264 spliced with medium-high-quality MPEG-2.]{
		\includegraphics[width=0.486\textwidth]{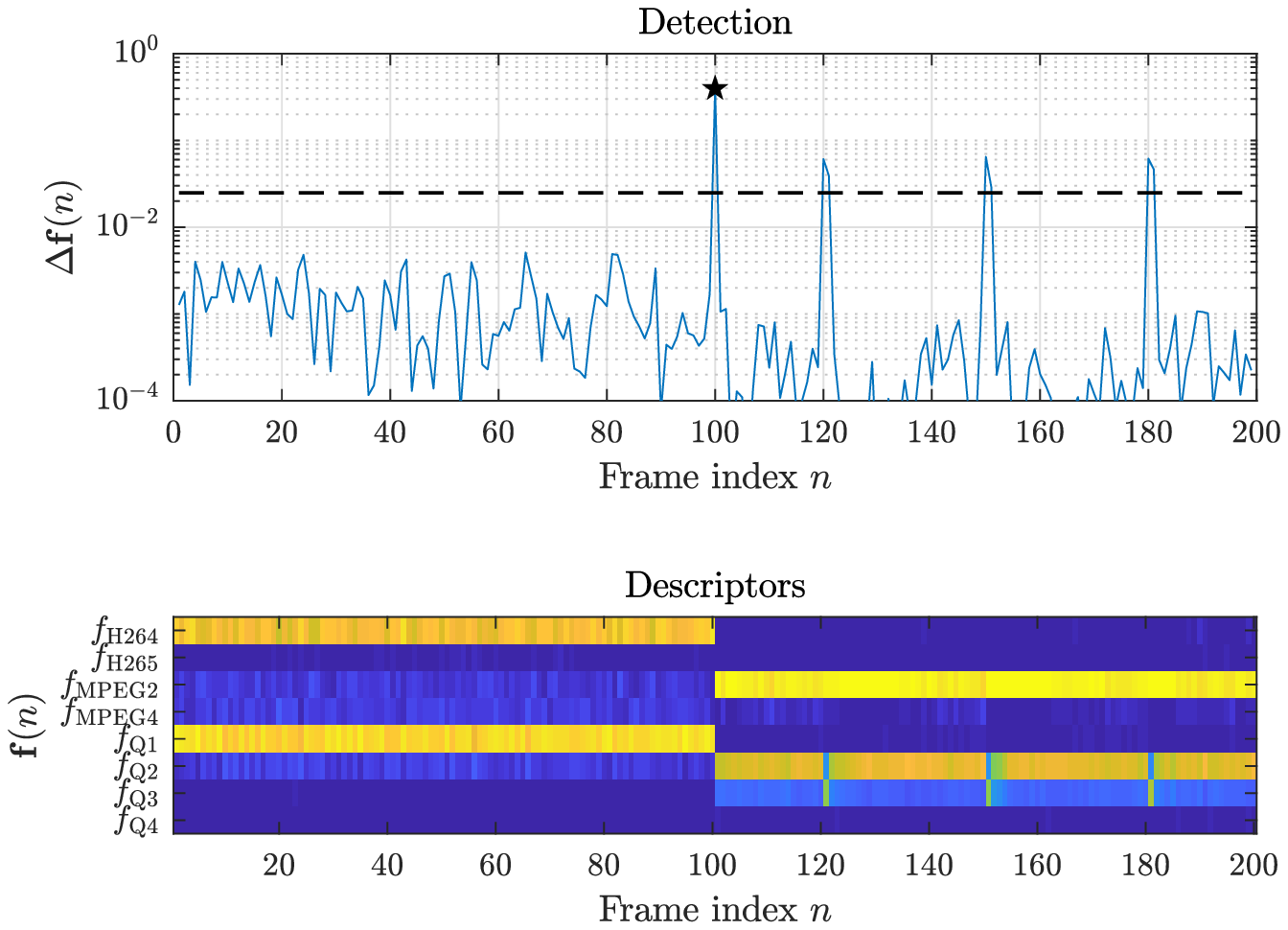}
		\label{fig:splicing_gop}}
	
\caption{Two examples of temporal splicing localization. Feature descriptors (below) are analyzed by means of the Euclidean distance (above) between adjacent vectors.
Figure \ref{fig:splicing_gop} also shows the presence of false positives due to intra-coded frames, with periodicity given by the \gls{gop} size.}\label{fig:splicing_detection}
\end{figure*}

This first implementation of \gls{focal} employs the \gls{cnn} described in \ref{sec:net} to extract from an input frame-patch a descriptor associated to its coding standard and quality. The idea is that patches or frames coming from different video sequences will exhibit different coding traces. Detecting descriptor inconsistencies may therefore lead to localizing forgeries.

The proposed \gls{cnn} was trained to solved a 4-class \emph{codec classification} task, within the following closed set of coding standards, $$\{\text{MPEG-2, MPEG-4, H.264, H.265}\},$$ and a 4-class \emph{quality classification} task, where the encoding quality is determined by the following values of the quantization step, $$\qs = \{5,10,20,40\}.$$ We will refer to the four quality levels throughout the paper as \emph{high}, \emph{medium-high}, \emph{medium-low} and \emph{low}, respectively.

The two trained models were kept separate and used as independent feature extractors (details on the training phase are provided in Sections \ref{sec:model_codec},\ref{sec:model_quality}. Note also that this framework is scalable to an arbitrary number of trained models, given that they output a vector-shaped feature descriptor.

In the following paragraphs, we discuss the feature extraction phase and the algorithms designed to detect inconsistencies in the descriptors, with the purpose of solving two typical video forensics scenarios: temporal and spatial splicing localization.

\subsection{Feature extraction}\label{sec:features}

Let $\X$ be a video sequence of $N$ frames, as defined in Section \ref{sec:problem}. Each frame $\X_n$ is split into $64 \times 64$ patches $\X^p_n$, $p \in [1, P]$, where the number of patches $P$ depends on the video resolution and on the stride used for patch extraction. Note that, in the case of overlapping patches, the stride must be a multiple of the dimension of the coding blocks (8 pixels), in order to maintain the alignment described in Section \ref{sec:net}. The extracted patches are then converted to YCbCr color space and their luma components are fed into the two trained \glspl{cnn}.

For each patch $\X^p_n$, the output of each network is a four-element feature vector,
\[
\arraycolsep = 0.2em
\begin{array}{lccr}
\fc^p(n)  =  [ f_\text{H264}^p(n), &  f_\text{H265}^p(n), & f_\text{MPEG2}^p(n), & f_\text{MPEG4}^p(n) ],\\
\\
\fq^p(n)  =  [ f_\text{low}^p(n), & f_\text{m-low}^p(n), & f_\text{m-high}^p(n), & f_\text{high}^p(n) ],
\end{array}
\]
where each element $f_{\{\cdot\}}^p(n)$ represents the likelihood of the $p$-th patch from the $n$-th frame being encoded with one of the four considered codecs/qualities. Due to the final softmax activation, feature vectors are non-negative and sum to one. 
As well as considering these vectors as probability distributions over codec/quality classes, one can interpret them as general descriptors capturing local coding traces. As a matter of fact, for forgery detection we are not required to exactly detect the adopted codec and the related coding parameters, but rather observe some sort of feature inconsistency between and within frames.

Given our patch-level descriptors, obtained from heterogeneous feature extractors, we can design different algorithms that leverage such information to detect anomalies, which in turn raise an alarm on the possible presence of forgeries.

\subsection{Temporal splicing localization}\label{sec:splicing}

The proposed temporal splicing localization algorithm relies on the presented features to calculate a descriptor for each frame of the video, and then look for inconsistencies between adjacent descriptors. Without loss of generality, we considered temporally-spliced videos composed by only two shots, since this can be easily extended by iterating the same procedure. Additionally, we considered the case of spliced videos obtained with sequences encoded with different codecs \mbox{and/or} different quality parameters, thus simulating the case of compilations of shots coming from different devices, broadcasting sources and social media, as well as shots compressed multiple times or re-encoded as a whole after being spliced. 

Given the patch-level features obtained with the procedure in \ref{sec:features}, the desired frame-level feature vectors, $\fc(n),\fq(n)$, are obtained through a standard average,
\begin{align}
\fc(n) & = \frac{1}{P} \sum_{p=1}^{P} \fc^p(n),\\
\fq(n) & = \frac{1}{P} \sum_{p=1}^{P} \fq^p(n),
\end{align}
where all operations are performed element-wise. 

Finally, the two vectors are concatenated into a general eight-element frame descriptor,
\begin{equation}
\fcq(n) = [\fc(n),\,\fq(n)].
\end{equation}

To automatically detect inconsistencies over time, we compute the squared Euclidean distance between adjacent feature vectors,
\begin{equation}
	\Delta\fcq(n) = \lVert \fcq(n) - \fcq(n+1)\rVert^2,
\end{equation}
and we feed $\Delta\fcq(n)$ to a threshold-detector. 

Figure \ref{fig:splicing_detection} reports two examples of temporal splicing localization, applied to 200-frame videos with a splicing point at frame $n=100$. 
Figure \ref{fig:splicing_ex} shows a 100-frame MPEG-2 medium-low-quality video, spliced with a 100-frame MPEG-4 low-quality video. We can observe an evident feature inconsistency at the splicing point, which is correctly detected in the Euclidean distance domain. Note that the $\Delta\fcq$ axis is displayed in logarithmic scale and the splicing peak is actually two orders of magnitude higher than the background.
Figure \ref{fig:splicing_gop} highlights the sensitivity of the algorithm to intra-coded frames. In the second half of the compilation, the system detects a strong inconsistency once every 30 frames (the \gls{gop} size), producing a series of false positives (even though the actual splicing point still yields a significantly higher peak). However, the pattern is regular by its very nature, and thus easy to neglect automatically. Interestingly, note how such inconsistencies are detected only by quality features (lower four), while codec ones remain, correctly, idle: the codec itself is not changing in presence of an I frame, but coding parameters are.

\subsection{Spatial splicing localization}\label{sec:tampering}

\begin{figure*}[t]
	\centering
	\subfigure[t][Feature tensor $\Fcq$ (classifier output).]{
		\includegraphics[width=0.48\textwidth]{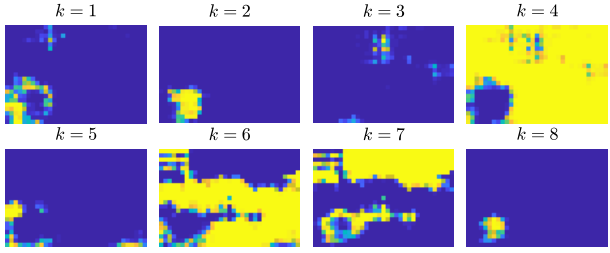}
		\label{fig:feat_tens}}%
	\subfigure[t][Activation tensor $\Hcq$.]{
		\includegraphics[width=0.48\textwidth]{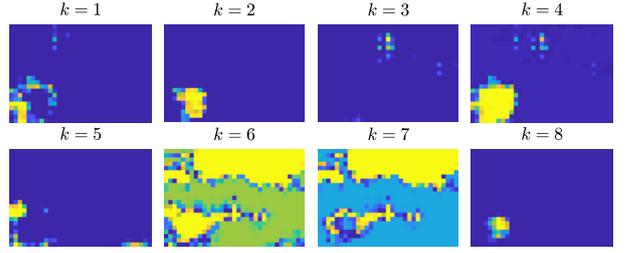}
		\label{fig:activ_tens}} \hfill
	\subfigure[t][\Acrfullpl{ver}.]{
		\includegraphics[width=0.486\textwidth]{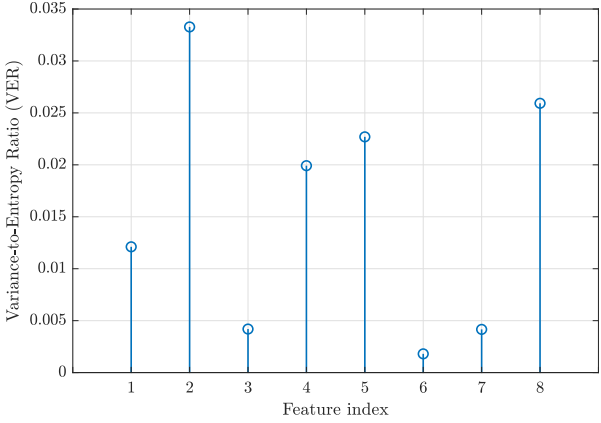}
		\label{fig:ver}}%
	\subfigure[t][Activation map fusion $\barHcq$.]{
		\includegraphics[width=0.486\textwidth]{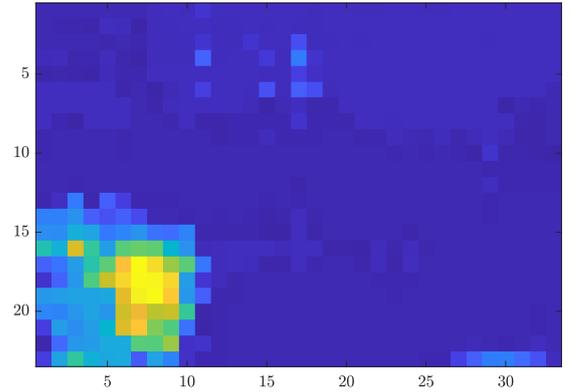}
		\label{fig:activ_map}}		
	\caption{Example of spatial splicing localization. The output maps of the \gls{cnn} classifiers in \ref{fig:feat_tens} are processed by the activation function in \eqref{eq:activ_func}, obtaining the activation maps in \ref{fig:activ_tens}. The latter are then averaged using the \glspl{ver} defined in \eqref{eq:ver} as weights, providing the final activation map in \ref{fig:activ_map}.}
	\label{fig:tampering_detection}
\end{figure*}

Differently from temporal forgeries, the problem of identifying and localizing a spatial splicing has to be solved within a single frame.
The proposed spatial splicing localization algorithm relies on the presented features to calculate local descriptors within the frame, and then look for possible coding inconsistencies. The presence of altered regions comes in the form of activation maps, containing the likelihood of being altered for each patch in the frame. Note how this scenario is significantly more challenging than temporal forgeries: having not the opportunity of averaging feature vectors, as in the case of frame-level descriptors, translates into reasonably less accurate features, and thus requires a more sophisticated processing.

Given a frame $\X_n$, patch-level descriptors $\fc^p(n)$ and $\fq^p(n)$ are extracted as described in \ref{sec:features}. We recommend an 8-pixel stride to have a dense description of the frame, while remaining aligned with the quantization grid.
The two vectors are concatenated into a general eight-element patch descriptor,
\begin{equation}
\fcq^p=[\fc^p(n),\,\fq^p(n)].
\end{equation}
Let $P_U,P_V$ be the number of patches extracted along the $U,V$ dimensions of frame $\X_n$, respectively. The patch-level descriptors are then arranged in a $P_U\times P_V\times 8$ frame-level \emph{feature tensor},
\begin{equation}
\Fcq(n)=
\begin{bmatrix}
\fcq^{1,1}(n) & \ldots & \fcq^{1,P_V}(n)\\
\vdots & \ddots & \vdots\\
\fcq^{P_U,1}(n) & \ldots & \fcq^{P_U,P_V}(n)
\end{bmatrix}.
\end{equation}
We call a \emph{feature map}, $\Fcq_k(n),\,k \in [1,8]$, each $P_U\times P_V \times 1$ matrix in the tensor. 

Figure \ref{fig:feat_tens} shows an example of feature tensor, with the eight feature maps plotted separately. In this example, the splicing is located at the bottom-left corner of the frame.

The second step consists in transforming the feature tensor into an \emph{activation tensor}, i.e., a set of eight activation maps. Each map $\Fcq_k(n)$ of the feature tensor is fed separately into a suitable activation function $h(\cdot)$ to produce an activation map $\Hcq_k(n)$. With the purpose of highlighting regions that differs from the general trend, we chose as activation function the pixel-wise squared distance from the average value,
\begin{equation}\label{eq:activ_func}
\Hcq_k(n) = h(\Fcq_k(n)) = \lVert \Fcq_k(n) - E\left(\Fcq_k(n)\right) \rVert^2, 
\end{equation}
where $E(\cdot)$ denotes the expectation operator.

Figure \ref{fig:activ_tens} shows the activation tensor obtained from the feature tensor in Figure \ref{fig:feat_tens}. Note the importance of the activation function for the feature map $k=4$, in particular.  

The last and most delicate step consists in fusing the activation tensor into the final activation map. The main issue is that not all feature maps are equally informative, in general: since each map activates in presence of a specific coding trace, it follows that maps carrying the highest information content will be those related to the coding parameters closest to what is actually present in the analyzed frame. In Figure \ref{fig:activ_tens}, for instance, we observe that: (i) maps number 1, 2, 4, 5 and 8 correctly agree on an activation at the bottom-left corner; (ii) map number 3 does not activate at all; (iii) maps number 6 and 7 show a noisy and wide activation, probably due to a background-foreground coding difference in the original video.

To automatically select the most informative maps, we devised a twofold criterion accounting for the possible presence of idle and/or widely-activated maps.
\begin{enumerate}
\item \emph{High variance}: a useful map should contain diversity in its values, if an activation is present. This condition helps filtering out idle maps.
\item \emph{Low entropy}: a useful activation should be localized to the tampered region. This condition helps filtering out noisy or widely-activated maps. 
\end{enumerate}

With these two conditions in mind, we defined a metric called \emph{\acrfull{ver}},
\begin{equation}\label{eq:ver}
\ver(x) = \frac{\text{var}(x)}{H(x)},\quad x \text{ r.v.}
\end{equation}
which merges the high-variance and low-entropy conditions into a single scoring value.

Figure \ref{fig:ver} reports the \glspl{ver} obtained for the activation maps in the current example. Note that the highest \gls{ver} values are related to the maps resulting the most informative according to the criteria outlined above, i.e., $k=2,4,5,8$.

The final fused activation map $\barHcq(n)$ for frame $\X_n$ is obtained as
\begin{equation}
\barHcq(n) = \frac{\sum\limits_{k=1}^K\left(\ver\left(\Hcq_k(n)\right)\cdot \Hcq_k(n)\right)}{\sum\limits_{k=1}^K \ver\left(\Hcq_k(n)\right)} ,
\end{equation}
which denotes an element-wise weighted average of the eight activation maps, with weights equal to the \glspl{ver}.

Figure \ref{fig:activ_map} shows the fusion result for the activation maps of the current example. As required, all noisy and flat activation maps are discarded, while meaningful ones are retained and merged together into a human-readable output.

%% file: results.tex
\section{Experiments and results}\label{sec:results}

A distinctive trait of \gls{focal} framework consists in training the same \gls{cnn} described in Section \ref{sec:net} to solve different classification tasks. In this work, we trained two independent coding-related models:
\begin{itemize}
\item a 4-class codec classifier; 
\item a 4-class quality classifier. 
\end{itemize}

This section describes the training of the proposed \gls{cnn} and the testing campaign carried out to evaluate the system in realistic scenarios, along with the obtained experimental results. Subsections \ref{sec:model_codec} and \ref{sec:model_quality} report the training details for the two employed models; \ref{sec:dataset} describes the generation of the testing dataset; \ref{sec:experiments_synth} and \ref{sec:experiments_real} outline the performed experiments in controlled and uncontrolled scenarios.

\subsection{Codec-CNN training}\label{sec:model_codec}

The model was trained on four video codecs: H.264, H.265, MPEG-2 and MPEG-4. 

A training dataset of 300 high-resolution videos was built, starting from five uncompressed video sequences: \emph{duckstakeoff} (720p), \emph{stockholm} (720p), \emph{ice} (4CIF), \emph{harbour} (4CIF), \emph{parkrun} (720p). Each video was encoded with the FFmpeg library to obtain 60 different versions, combining the four codecs with different coding configurations: fixed quality parameter $\qp$ ranging from 1 to 10; constant bitrate set to 2 Mb/s, 4 Mb/s and 6 Mb/s; variable bitrate set to 2 Mb/s, 4 Mb/s and 6 Mb/s; \gls{gop} of 30 frames.

To validate the codec classification network, we built a similar dataset of 300 high-resolution videos following the same procedure adopted for the training phase, but starting from a different set of original video sequences: \emph{parkjoy} (720p), \emph{parkrun} (720p), \emph{shields} (720p), \emph{soccer} (4CIF), and \emph{stockholm} (720p).

To test the the codec classification network on a completely unrelated set, we built a dataset of 1672 low-resolution videos, starting from 19 sequences at CIF resolution: \emph{akiyo}, \emph{crew}, \emph{mother}, \emph{soccer}, \emph{bridgeclose}, \emph{flower}, \emph{news}, \emph{table}, \emph{city}, \emph{foreman}, \emph{paris}, \emph{tempete}, \emph{coastguard}, \emph{hall}, \emph{salesman}, \emph{waterfall}, \emph{container}, \emph{mobile},
\emph{signirene}. Each video was encoded with FFmpeg mixing codecs and qualities: fixed quality parameter $\qp$ ranging from 1 to 31 with step 2; constant bitrate set to 500 Kb/s, 1 Mb/s and 2 Mb/s; variable bitrate set to 500 Kb/s, 1 Mb/s and 2 Mb/s; \gls{gop} of 10 frames.

The network was trained with the following parameters: Adam optimizer with standard parameters and learning rate, categorical cross-entropy loss function. We selected the model minimizing the validation loss over 50 epochs.

\subsection{Quality-CNN training}\label{sec:model_quality} 
The model was trained on four quality levels identified by the quantization step $\qs$. Such value is not directly accessible in the majority of codecs, as it is usually controlled by a higher-level quality parameter $\qp$, the implementation of which may differ from codec to codec. As a matter of fact, the relation between $\qs$ and $\qp$ is exponential in H.264 \eqref{eq:qs-h26x}, and piece-wise linear in MPEG-2 and MPEG-4 \eqref{eq:qs-mpegx}:
\begin{align}
& \qs_{\mathrm{H264}}  	& = & \quad\frac{5}{8}\cdot 2^{\displaystyle\nicefrac{\qp}{6}},\label{eq:qs-h26x}\\
& \qs_{\mathrm{MPEG}} 	& = &
\quad\begin{cases}
8, & 1\leq\qp\leq 4\\
2\qp, & 5\leq\qp\leq 8\\
\qp+8, & 9\leq\qp\leq 24\\
2\qp-16, & 25\leq\qp\leq 31\\
\end{cases}.\label{eq:qs-mpegx}
\end{align} 

To generate the training, validation and testing sets, we considered three different codecs, namely MPEG-2, MPEG-4 and H.264, and tuned the respective quality parameters according to \eqref{eq:qs-h26x} and \eqref{eq:qs-mpegx} in order to have the same quantization step.

Seven raw videos in YUV format and 4CIF quality were used: \emph{crew}, \emph{crowdrun}, \emph{duckstakeoff}, \emph{harbour}, \emph{ice}, \emph{parkjoy}, \emph{soccer}. Each video was encoded with FFmpeg, using three codecs (MPEG-2, MPEG-4, H.264), four quantization steps ($\qs=\{5,10,20,40\}$), variable bitrate (VBR) and \gls{gop} 30, yielding a total of 84 video sequences.
From each video, we extracted 30 frames and 99 non-overlapping $64\times64$ patches per frame, yielding a total of 249480 patches. However, we observed a clear performance improvement, both in training and testing, using only high-variance patches, since ``flat'' ones tend to look alike in every codec-quality configuration. We set a variance threshold of $10^3$, ending up retaining 122473 patches (about 50\% of the total). The set of patches was partitioned as follows: 70\% for training, 20\% for validation and 10\% for testing.

The network was trained with the following parameters: \gls{sgdm} optimizer, categorical cross-entropy loss function, initial learning rate $5\cdot10^{-3}$ with drop factor 0.5 every 5 epochs, batch size 256. We selected the model minimizing the validation loss over 50 epochs.

\subsection{Testing dataset}\label{sec:dataset}

The generation process of the final testing dataset was split into two steps. First, we created a set $\D$ of encoded videos, using different codecs and qualities. Then, we used the videos in $\D$ to produce two datasets: one set $\Dtemp$ of temporally-spliced videos and one set $\Dspat$ of spatially-spliced videos.

Dataset $\D$ was generated starting from five uncompressed videos that were not included in the previous sets used for training, validation and testing: \emph{four people}, \emph{in to tree}, \emph{johnny}, \emph{kristen and sara}, \emph{old town cross}. Each video is 210 frames long and 720p resolution. Using FFmpeg, each sequence was encoded with MPEG-2, MPEG-4 and H.264 codecs, with \gls{gop} set to 30 frames and four fixed values of the quality parameter, $\qp=\{3,8,13,18\}$, for a total of 12 different versions of the same video. Overall, the dataset consisted of 60 encoded videos, or 12600 frames. The choice of using $\qp$ for the final system evaluation is motivated by two reasons: first, it allows to test the algorithm in a scenario closer to a real-world case, since in everyday applications the encoding quality is tuned by means of the quality parameter, not the quantization step directly; then, it produces a testing set that is even more uncorrelated with that used in the \gls{cnn} training and testing phases.

Denoting with $v_i$, $i\in[1,5]$ a video from the five originals, dataset $\D$ consists of $$\D = \{\D_{v_1}, \ldots, \D_{v_5} \},$$ where $\D_{v_i}\subset\D$, $|\D_{v_i}|=12$, is the subset containing all the different versions of $v_i$.

Dataset $\Dtemp$ for temporal splicing localization was obtained by splicing the first 100 frames of each video in $\D_{v_i}$ with the second 100 frames of any other video in $\D_{v_i}$, for each $i$. Given that the number of possible pairs in a set of 12 elements is $\binom{12}{2}=66$, dataset $\Dtemp$ consists of $5\times 66=330$ temporally-spliced videos, corresponding to $330\times 200 = 66000$ frames.

Dataset $\Dspat$ for spatial splicing localization was obtained by substituting a CIF window ($288\times 352$) of each video in $\D_{v_i}$ with the same window of any other video in $\D_{v_i}$, for each $i$. The window was placed at the center of the frame, with the top-left corner aligned with the patch extraction grid, and kept fixed throughout the length of the video. Similarly to $\Dtemp$, dataset $\Dspat$ consists of 330 spatially-spliced videos, corresponding to 66000 frames, or 1452000 patches.

To simulate a more realistic scenario, videos in $\Dtemp$ and $\Dspat$ were re-encoded with high-quality H.264 after the forgery. Note that, since we are using different versions of the same video, there are no scene changes or content inconsistencies in the forged sequences: temporally-spliced videos have the first 100 frames encoded differently from the subsequent ones; spatially-spliced videos have a CIF window in the middle of the frame encoded differently from the rest. This is necessary to assess the capability of our algorithm to properly localize changes in coding rather than content.

\subsection{Experiments in controlled environment}\label{sec:experiments_synth}

For each forensic scenario, namely temporal and spatial splicing, we run three separate tests: one using codec-related features alone; one using quality-related features alone; one using the concatenated features.

Dataset $\Dtemp$ was analyzed with the algorithm outlined in Section \ref{sec:splicing}. The detection of splicing points was evaluated frame-wise: a true-positive consisted of a splicing point correctly identified in the transition between two adjacent frames. For this experiment, false positives related to the \glspl{gop} were discarded, as discussed in Section \ref{sec:splicing}.

\newcommand{\rewsize}{0.159}
\begin{table*}[t]
\setlength{\tabcolsep}{2pt}
	\centering
	\caption{Forgery localization at different re-encoding qualities}
	\begin{threeparttable}
		\begin{tabular}{cccccc}
			\toprule
			Original frame & Forged frame & Lossless & $q=10$ & $q=20$ & $q=30$\\
			\midrule
			\includegraphics[width=\rewsize\textwidth]{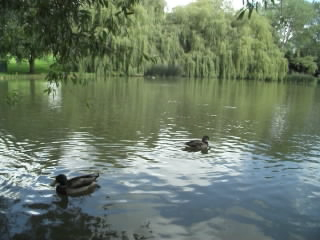} & \includegraphics[width=\rewsize\textwidth]{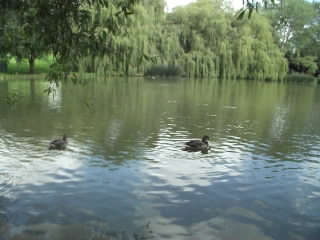} & \includegraphics[width=\rewsize\textwidth]{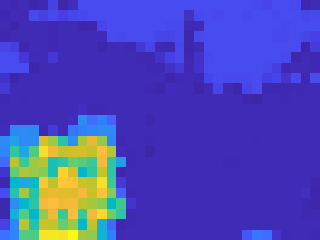} & \includegraphics[width=\rewsize\textwidth]{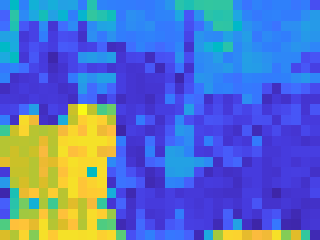} & \includegraphics[width=\rewsize\textwidth]{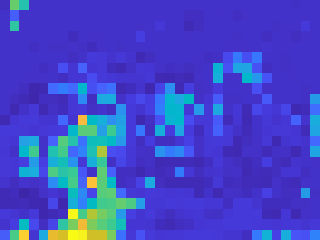} & \includegraphics[width=\rewsize\textwidth]{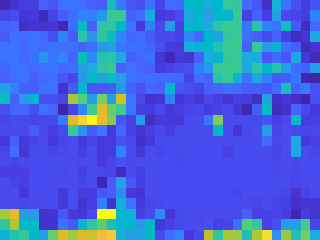}\\
			\midrule
			\includegraphics[width=\rewsize\textwidth]{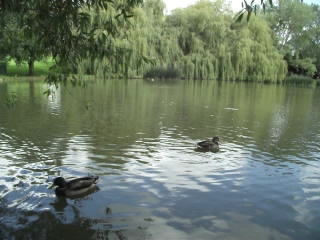} & \includegraphics[width=\rewsize\textwidth]{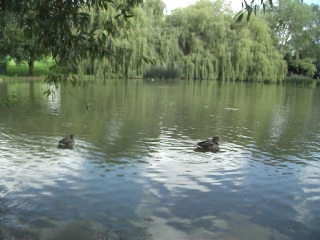} & \includegraphics[width=\rewsize\textwidth]{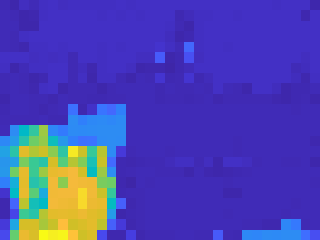} & \includegraphics[width=\rewsize\textwidth]{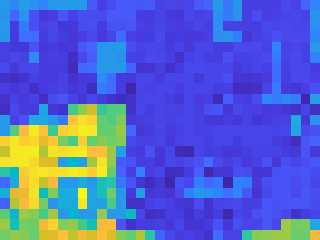} & \includegraphics[width=\rewsize\textwidth]{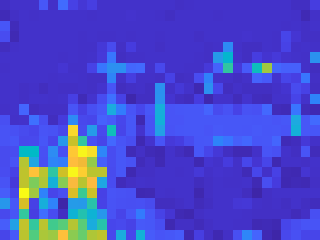} & \includegraphics[width=\rewsize\textwidth]{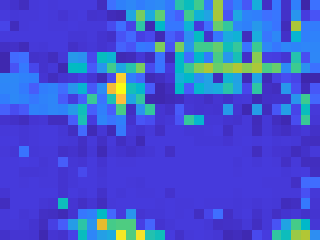}\\
			\midrule
			\includegraphics[width=\rewsize\textwidth]{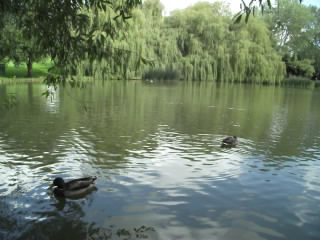} & \includegraphics[width=\rewsize\textwidth]{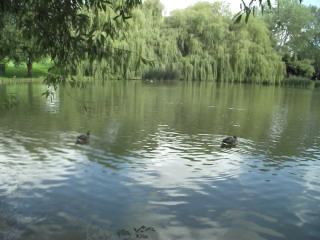} & \includegraphics[width=\rewsize\textwidth]{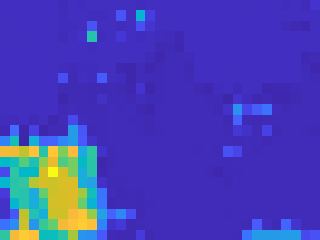} & \includegraphics[width=\rewsize\textwidth]{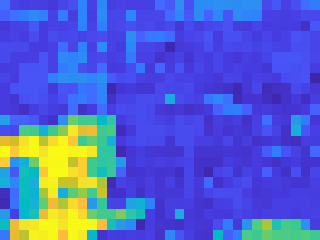} & \includegraphics[width=\rewsize\textwidth]{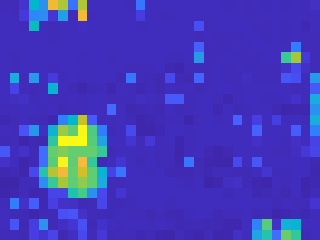} & \includegraphics[width=\rewsize\textwidth]{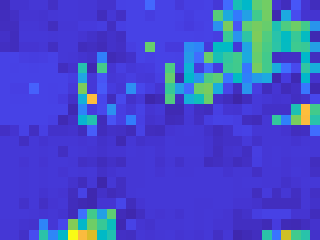}\\
			\bottomrule 
		\end{tabular}
		\begin{tablenotes}
			\item Decaying localization performance at lower re-encoding qualities of the same forged video. From top to bottom, frames 52, 78 and 144 of sequence \emph{01} of the \acrshort{rewind} dataset \cite{Bes13:video_tamp_res}. From left to right, the original frame, the forged frame, the localization heatmaps for lossless and lossy H.264 encoding, with quality parameter $q=10,20,30$.
		\end{tablenotes}
	\end{threeparttable}
\label{tab:rewind_qualities}
\end{table*}

\begin{figure}
\centering
\includegraphics[width=1\columnwidth]{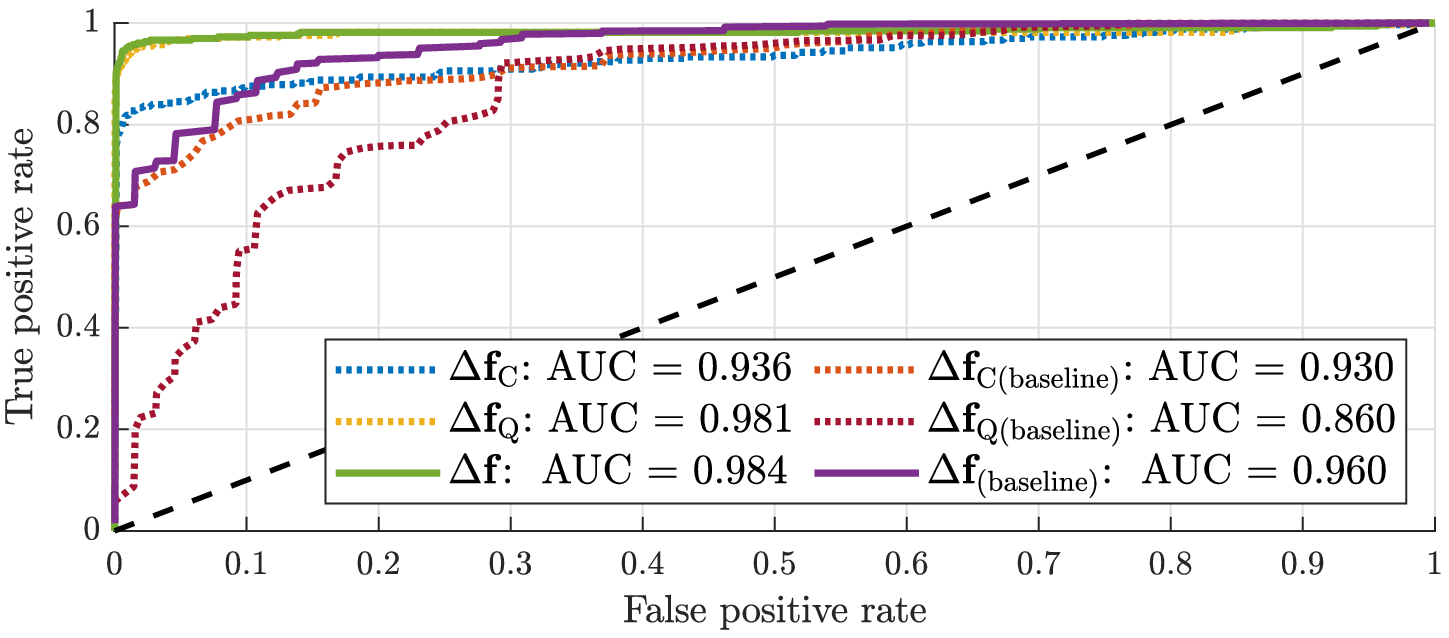}
\caption{\Gls{roc} curves for frame-wise temporal splicing localization. Comparison of codec-related, quality-related and combined features, for the proposed method and the baseline \cite{Ver18:video_splicing}.}\label{fig:roc_splicing}
\centering
\includegraphics[width=1\columnwidth]{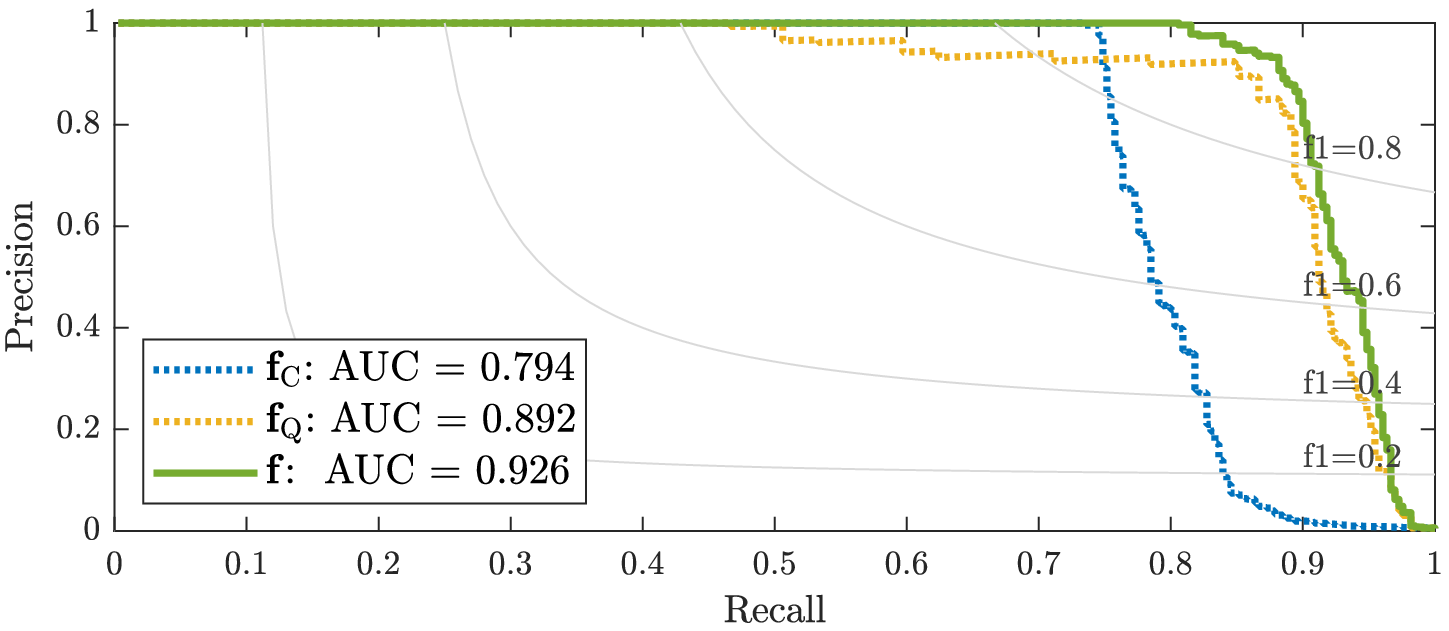}
\caption{\gls{pr} curves for frame-wise temporal splicing localization. Comparison of codec-related, quality-related and combined features.}\label{fig:prc_splicing}
\end{figure}

\begin{figure}
\centering
\includegraphics[width=1\columnwidth]{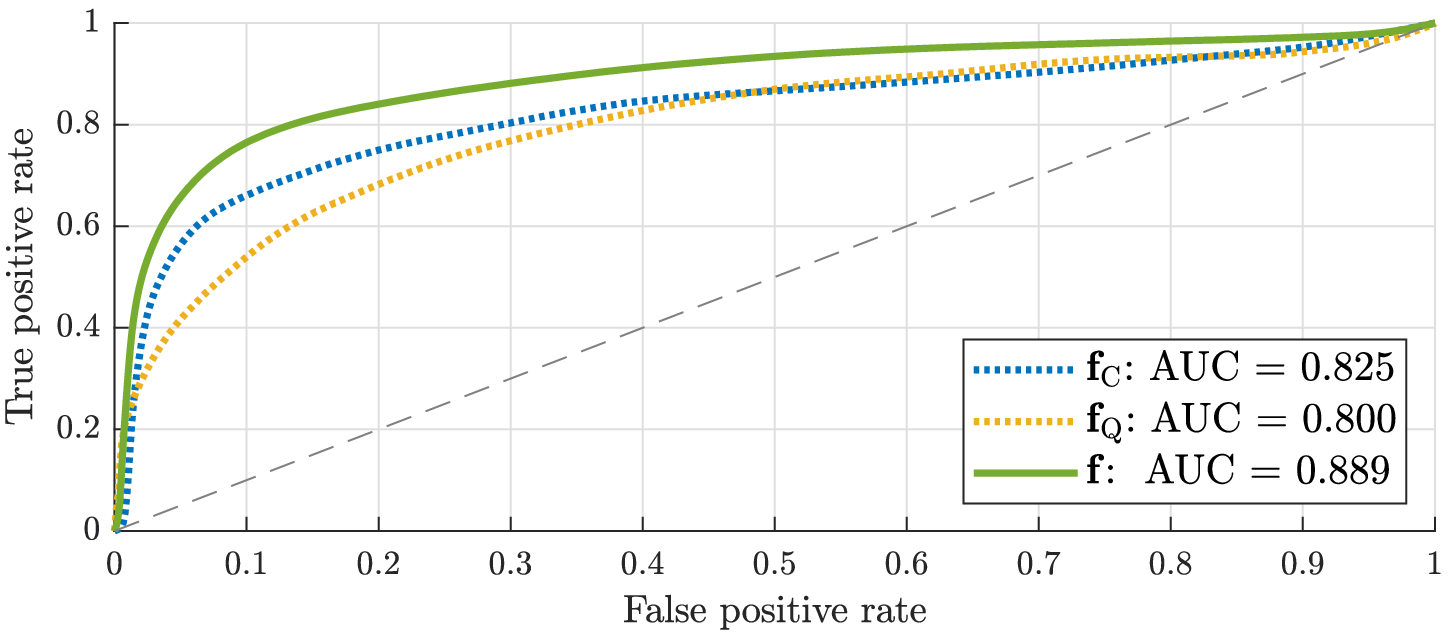}
\caption{\gls{roc} curves for patch-wise spatial splicing localization, using descriptors from a \emph{single} frame. Comparison of codec-related, quality-related and combined features.}
\label{fig:roc_tampering_single}
\vspace{5ex}
\centering
\includegraphics[width=1\columnwidth]{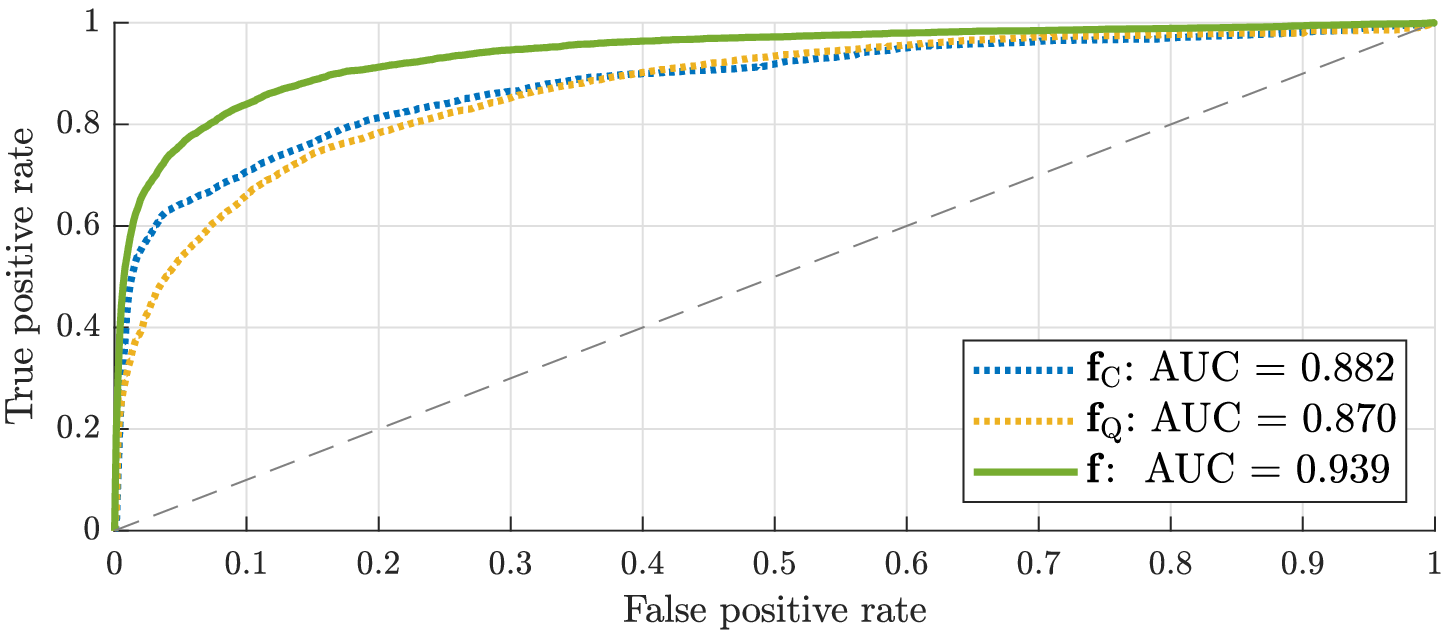}
\caption{\gls{roc} curves for patch-wise spatial splicing localization, using descriptors averaged over \emph{multiple} frames. Comparison of codec-related, quality-related and combined features.}
\label{fig:roc_tampering_mult}
\end{figure}

Figure \ref{fig:roc_splicing} reports the \gls{roc} curves obtained for temporal splicing localization with the three different features. For each case, a direct comparison with the baseline work in \cite{Ver18:video_splicing} is provided. All three descriptors outperform the baseline ones, with an \gls{auc} peaking at 0.984 for the concatenated descriptor. For better appreciating the small differences between concatenated features and separate ones in the proposed method (that appear squished in the top-left corner), we report the same results displayed as \gls{pr} curves in Figure \ref{fig:prc_splicing}.
We can observe how quality-related features $\fq$ provide better results than codec-related ones $\fc$ in this scenario. However, the concatenated features $\fcq$ still lead to an improvement with respect to $\fq$ alone. Tests run on the concatenated descriptor $\fcq$ show a 100\% precision up to a recall of roughly 80\%, denoting a good robustness of the algorithm to false positives, and an \gls{auc} of $0.926$. The optimal operating point of the green curve corresponds to an F1-measure of $0.902$.

Dataset $\Dspat$ was analyzed with the algorithm outlined in Section \ref{sec:tampering}. The detection was evaluated patch-wise: a true-positive consisted of a $64\times 64$ patch correctly classified as forged. We run two separate tests: one with patch-level descriptors obtained frame-by-frame; one with descriptors obtained by averaging throughout the video frames.

\newcommand{\picsize}{0.153}
\begin{table}
	\setlength{\tabcolsep}{2pt}
	\centering
	\caption{Forgery localization examples}
	\begin{threeparttable}
		\begin{tabular}{ccc}
			\toprule
			Original frame & Forged frame & Detection\\
			\midrule
			\includegraphics[width=\picsize\textwidth]{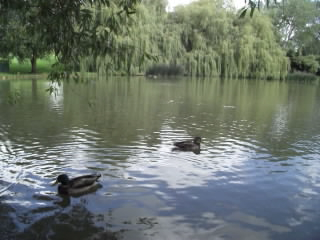} & \includegraphics[width=\picsize\textwidth]{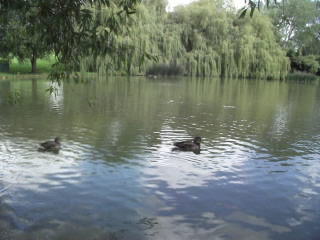} & \includegraphics[width=\picsize\textwidth]{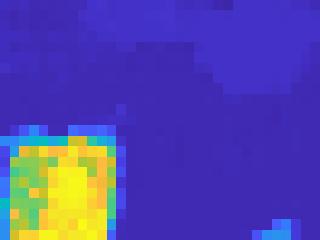}\\
			\midrule
			\includegraphics[width=\picsize\textwidth]{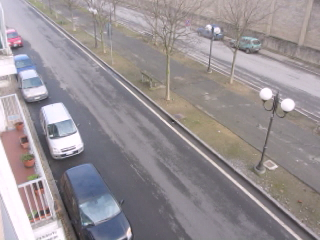} & \includegraphics[width=\picsize\textwidth]{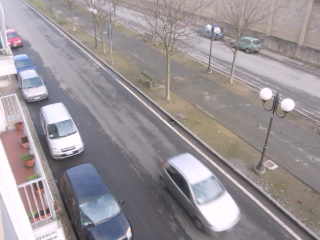} & \includegraphics[width=\picsize\textwidth]{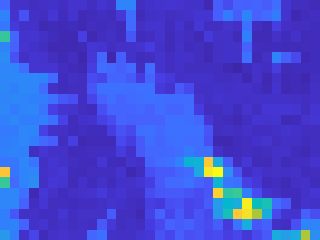}\\
			\midrule
\includegraphics[width=\picsize\textwidth]{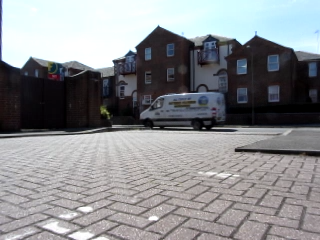} & \includegraphics[width=\picsize\textwidth]{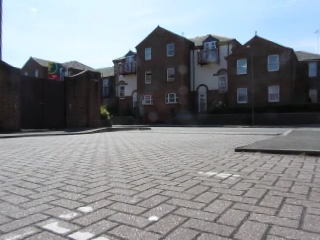} & \includegraphics[width=\picsize\textwidth]{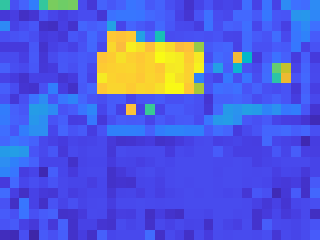}\\
			\bottomrule 
		\end{tabular}
		\begin{tablenotes}
			\item Example frames from a sample of video sequences in the \gls{rewind} dataset \cite{Bes13:video_tamp_res}.
		\end{tablenotes}
	\end{threeparttable}
\label{tab:rewind_examples}
\end{table}

Figure \ref{fig:roc_tampering_single} reports the \gls{roc} curves obtained for spatial splicing localization on a single frame, with the three different features. Again, the benefits of using concatenated descriptors are clearly visible. Note also how in this case $\fc$ performs better than $\fq$, when taken individually. We can assume this is due to the fact that quality is trickier to assess at patch-level than at frame-level: low-variance patches, for instance, typically look very similar at different encoding qualities. However, $\fq$ features still provide useful information in combination with
$\fc$, as shown by the results improvement associated with $\fcq$.

Figure \ref{fig:roc_tampering_mult} shows the \gls{roc} curves obtained for spatial splicing localization averaged over multiple frames, with the three different features. As expected, we observe a clear improvement in all descriptors with respect to the single-frame case. Since in $\Dspat$ the forged region is fixed in time, we were able to average descriptors throughout all the 200 frames of the video sequences. In a real-case scenario, the assumption of a non-moving forged region does not hold, in general. However, it is still possible to resort to this performance-enhancing strategy by averaging descriptors over short-time windows, assuming that the motion of the altered region is slow enough, compared to the frame-rate.

All three experiments show that concatenating feature descriptors associated to different classification tasks lead to a clear performance improvement. As long as we are able to identify additional classes of forensic traces, we can assume that the proposed framework would keep taking advantage from a higher number of trained models.

\subsection{Experiments on uncontrolled videos}\label{sec:experiments_real}

As a last experiment, we run \gls{focal} on videos from the online dataset of the \gls{rewind} project \cite{Bes13:video_tamp_res}. These sequences contain photo-realistic forgeries, similar to those encountered in a practical forensic scenario. Original videos were recorded using low-end devices, with a resolution of $320\times240$ pixels and a framerate of 30 fps. Each forged sequence is available in four encoding configurations: lossless H.264 and lossy H.264 with $q=10,20,30$. 

Given the availability of multiple coding qualities for the same forged video, we used this dataset to assess how the robustness of the algorithm is impaired as the encoding quality decreases. 
Table \ref{tab:rewind_qualities} shows the activation maps calculated by \gls{focal} on three frames of sequence \emph{01} of the \gls{rewind} dataset. Each map was obtained from a single frame, with no temporal averaging. In this example, the forgery consists in a spatial splicing at the bottom-left corner of the frame: the bigger duck is spliced-out by copy-moving an empty portion of the water surface and a second duck is spliced-in. The resulting frame still appears photo-realistic and spotting the forgery turns out being a challenging task even for a human observer. As we can see, the localization capabilities of the algorithm remain satisfying up to a quality parameter $q=10$. Lower encoding qualities progressively erase any useful forensic trace, impairing the detection reliability. However, since the content itself of the video become progressively less discernible, it is arguable whether a forger should ever adopt such low encoding qualities with the purpose of creating a convincing fake. 

Finally, Table \ref{tab:rewind_examples} reports a series of example frames from different videos of the \gls{rewind} dataset.

%% file: conclusion.tex

\section{Conclusions}\label{sec:conclusion}

In this paper we presented \gls{focal}, a framework for video forgery localization based on the self-consistency of coding traces. The main contributions that come along with this strategy consist of the design of an ad-hoc \gls{cnn} architecture for learning codec-related features and a fusion technique (using the proposed \acrlong{ver} metric) that merges different features into a general likelihood map, making the framework scalable and generalizable at will. A first implementation, featuring two independently-trained coding-related descriptors, was here proposed and tested over two different forgeries situations. Experimental results showed a clear performance improvement with respect to the previous work on temporal splicing localization, and promising results in the newly tackled scenario of spatial splicing localization.

Being able to capture local coding traces over small patches of a video-frame, paves
the way to several possibilities in the context of forgery detection. Future research will be devoted to assessing the scalability of the proposed framework through the addition of further models and by upgrading the existing ones with a higher number of coding parameters. New strategies to fuse and leverage feature information could be explored as well, with the purpose of enabling the detection of increasingly complex types of forgeries.